%% file: paper.tex
\RequirePackage[hyphens]{url}
\documentclass[runningheads]{llncs}

\usepackage[utf8]{inputenc} 
\usepackage[T1]{fontenc}    
\usepackage{url}            
\usepackage{booktabs}       
\usepackage{amsfonts}       
\usepackage{nicefrac}       
\usepackage{microtype}      
\usepackage{xcolor}         
\definecolor{codegreen}{rgb}{0,0.6,0}

\usepackage{epsfig}
\usepackage{endnotes}

\usepackage{graphicx}
\usepackage{array}
\usepackage{comment}

\usepackage{amsmath}
\usepackage{amssymb}
\usepackage{pifont}

\usepackage[accsupp]{axessibility}  

\usepackage[pagebackref,breaklinks,colorlinks]{hyperref}

\usepackage[utf8]{inputenc}
\usepackage[utf8]{luainputenc}

\usepackage{ifthen}

\usepackage{nicefrac}

\newif\ifcomments
\commentsfalse
\ifcomments
\usepackage[textsize=small,color=lightgray]{todonotes}
\else
\usepackage[disable]{todonotes}
\fi

\usepackage{textcomp}

\usepackage{pgfplots}
\pgfplotsset{compat=1.13}

\usepackage{enumitem}

\usepackage{etoolbox}
\makeatletter
\patchcmd{\ttlh@hang}{\parindent\z@}{\parindent\z@\leavevmode}{}{}
\patchcmd{\ttlh@hang}{\noindent}{}{}{}
\makeatother

\usepackage{balance}

\usepackage{datetime}

\usepackage{listings, listings-rust}

\usepackage[capitalize]{cleveref}
\crefname{section}{Sec.}{Secs.}
\Crefname{section}{Section}{Sections}
\Crefname{table}{Table}{Tables}
\crefname{table}{Tab.}{Tabs.}
\Crefname{listing}{Listing}{Listings}
\crefname{listing}{Lst.}{Lst.}
\AtBeginEnvironment{subappendices}{\crefalias{section}{appendix}}

\usepackage{xfrac}

\usepackage{rotating}
\usepackage{xparse}

\usepackage{outlines}

\usepackage{xspace}
\usepackage{framed}

\usepackage[page,toc]{appendix}
\usepackage{titletoc}

\usepackage{setspace}

\usepackage{adjustbox}

\newboolean{draft}
\setboolean{draft}{false}

\ifcomments
\paperwidth=\dimexpr \paperwidth + 7cm\relax
\oddsidemargin=\dimexpr\oddsidemargin + 4cm\relax
\evensidemargin=\dimexpr\evensidemargin + 4cm\relax
\marginparwidth=\dimexpr \marginparwidth + 4cm\relax
\fi

\newcommand{\ionel}[1]{\todo[color=red!40]{\textbf{Ionel}: #1}\ignorespaces}
\newcommand{\eyal}[1]{\todo[color=gray!40]{\textbf{Eyal}: #1}\ignorespaces}

\newcommand{\justin}[1]{\todo[color=cyan!40]{\textbf{Justin}: #1}\ignorespaces}

\newcommand{\myparagraph}[1]{\noindent{\bfseries #1.}}

\NewDocumentCommand{\rot}{O{45} O{1em} m}{\makebox[#2][l]{\rotatebox{#1}{#3}}}

\usepackage{comment}

\usepackage{color}

\ifdraft
 \usepackage{type1cm}
 \usepackage{eso-pic}
 \makeatletter
 \AddToShipoutPicture{%
             \setlength{\@tempdimb}{.5\paperwidth}%
             \setlength{\@tempdimc}{.53\paperheight}%
             \setlength{\unitlength}{1pt}%
             \put(\strip@pt\@tempdimb,\strip@pt\@tempdimc){%
         \makebox(0,700){\textcolor[gray]{0.50}%
         {\fontsize{15pt}{15pt}\selectfont{Draft of \today{}, {\currenttime} -- please do not distribute.}}}%
         \makebox(0,0){\rotatebox{45}{\textcolor[gray]{0.92}%
         {\fontsize{4cm}{4cm}\selectfont{DRAFT}}}}%
             }%
 }
\fi

\usepackage[capitalize]{cleveref}
\crefformat{section}{#2\S#1#3}

\DeclareMathAlphabet{\mathcal}{OMS}{cmsy}{m}{n}

\newcommand{\one}{({\em i}\/)}
\newcommand{\two}{({\em ii}\/)}
\newcommand{\three}{({\em iii}\/)}

\newcommand{\system}{Octopus}

\newcommand{\av}{AV}

\newcommand{\param}{metaparameter}
\newcommand{\Param}{Metaparameter}
\newcommand{\Params}{Metaparameters}
\newcommand{\params}{metaparameters}

\usepackage[labelfont=bf]{caption}
\usepackage[aboveskip=2pt,belowskip=1pt]{subcaption}

\newcommand{\thss}{\textsuperscript{th}}

\newcommand{\mota}{MOTA}
\newcommand{\motp}{MOTP}
\newcommand{\tmota}{S-MOTA}
\newcommand{\tmotp}{S-MOTP}
\newcommand{\tid}{S-ID}
\newcommand{\tfp}{S-FP}
\newcommand{\tfn}{S-FN}

\newlength{\oldtextfloatsep}
\newlength{\oldintextsep}

\usepackage{multirow}

\usepackage{makecell}

\newenvironment{tightitemize}{%
\begin{list}{$\bullet$}{%
\setlength{\itemsep}{1.5pt}%
\setlength{\topsep}{2pt}%
\setlength{\parskip}{0pt}%
\setlength{\parsep}{0pt}%

\setlength{\labelwidth}{0pt}%
\setlength{\leftmargin}{4pt}%
\setlength{\labelsep}{0pt}%
\setlength{\listparindent}{0pt}%
}}%
{\end{list}}

\newenvironment{tightenumerate}{%
\begin{list}{\labelenumi}{%
\usecounter{enumi}%
\setlength{\itemsep}{1.5pt}%
\setlength{\topsep}{2pt}%
\setlength{\parskip}{0pt}%
\setlength{\parsep}{0pt}%

\setlength{\labelwidth}{0pt}%
\setlength{\leftmargin}{4pt}%
\setlength{\labelsep}{0pt}%
\setlength{\listparindent}{0pt}%
}}%
{\end{list}}

\begin{document}

\pagestyle{headings}
\mainmatter
\def\ECCVSubNumber{2624}  

\title{Context-Aware Streaming Perception in Dynamic Environments}

\titlerunning{Context-Aware Streaming Perception}
\author{
  Gur-Eyal Sela\inst{1}\index{Sela, Gur-Eyal} \and
  Ionel Gog\index{Gog, Ionel}\inst{1}\thanks{Now at Google Research.} \and
  Justin Wong\index{Wong, Justin}\inst{1} \and
  Kumar Krishna Agrawal\index{Agrawal, Kumar Krishna}\inst{1} \and
  Xiangxi Mo\index{Mo, Xiangxi}\inst{1} \and
  Sukrit Kalra\index{Kalra, Sukrit}\inst{1} \and
  Peter Schafhalter\index{Schafhalter, Peter}\inst{1} \and
  Eric Leong\index{Leong, Eric}\inst{1} \and
  Xin Wang\index{Wang, Xin}\inst{2} \and
  Bharathan Balaji\index{Balaji, Bharathan}\inst{3}\thanks{Work unrelated to Amazon.} \and
  Joseph Gonzalez\index{Gonzalez, Joseph}\inst{1} \and
  Ion Stoica\index{Stoica, Ion}\inst{1}}
\authorrunning{G.-E. Sela et al.}
\institute{University of California, Berkeley \and Microsoft Research \and Amazon}

\maketitle

\thispagestyle{empty}

\input{abstract}

\input{introduction}


\input{related}
\input{problem-setup}

\input{methods}
\input{evaluation}

\input{conclusions}

\newpage
\bibliographystyle{splncs04}
\bibliography{paper}

\newpage

\setlength{\textfloatsep}{\oldtextfloatsep}
\setlength{\intextsep}{\oldintextsep}
\begin{appendix}
  \input{appendix.tex}
\end{appendix}

\end{document}

%% file: abstract.tex
\begin{abstract}
    Efficient vision works maximize accuracy under a latency budget. These
works evaluate accuracy offline, one image at a time. However, real-time vision
applications like autonomous driving operate in streaming settings, where
ground truth changes between inference start and finish. This results in a
significant accuracy drop. Therefore, a recent work proposed to maximize
accuracy in streaming settings on average. In this paper, we propose to
maximize streaming accuracy for every environment context. We posit that
scenario difficulty influences the initial (offline) accuracy difference, while
obstacle displacement in the scene affects the subsequent accuracy degradation.
Our method, Octopus, uses these scenario properties to select configurations
that maximize streaming accuracy at test time. Our method improves tracking
performance (\tmota{}) by $7.4\%$ over the conventional static approach.
Further, performance improvement using our method comes in addition to, and not
instead of, advances in offline accuracy.

\end{abstract}

%% file: introduction.tex
\section{Introduction}
\label{s:introduction}
\begin{figure*}[tb]
  \centering
  \captionsetup[subfigure]{width=0.99\textwidth}
  \begin{subfigure}{.49\textwidth}
    \centering
    \includegraphics[width=0.99\textwidth]{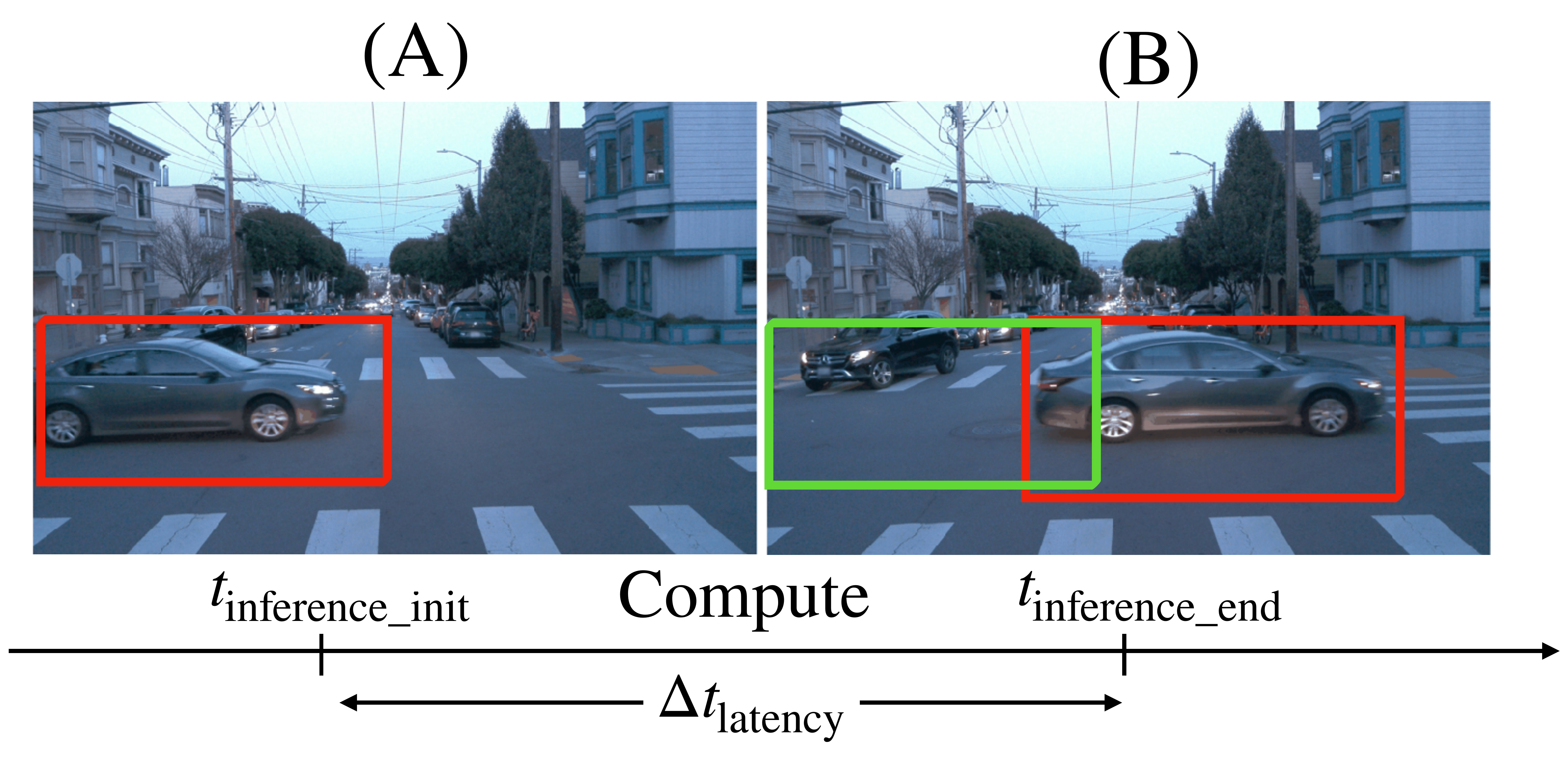}
    \caption{\small In online streaming settings, the environment
      changes during inference. Streaming accuracy is computed by evaluating
      the prediction run on ground truth A against ground truth B.}
    \label{f:tsp-figure}
  \end{subfigure}
  \begin{subfigure}{.49\textwidth}
    \centering
    \includegraphics[width=0.9\textwidth]{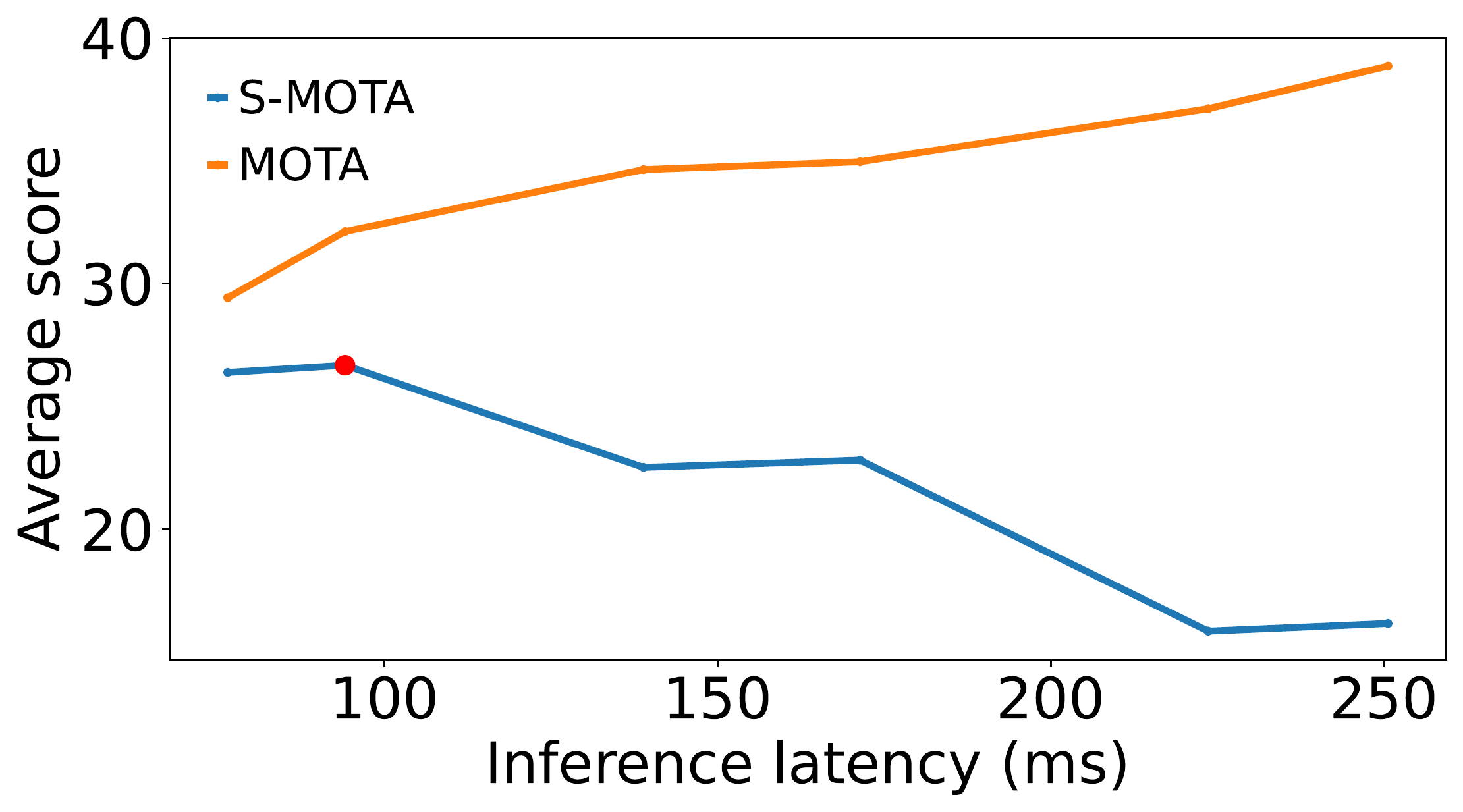}
    \caption{\small Offline and streaming accuracy of a tracker of
        increasing model size. While the offline \mota{} (orange, y-axis) of the
      model increases as inference latency (x-axis) increases, its streaming
      \mota{} (\tmota{} in blue, y-axis) decreases.}
    \label{f:t-mota-sync-mota}
  \end{subfigure}
  \caption{\small \textbf{Streaming accuracy deviates from offline accuracy because
    the ground truth changes during inference.}}
  \label{f:tradeoff-scenario}
\end{figure*}

Recent works like EfficientDet~\cite{tan20efficientdet}, YOLO~\cite{yolov4}, and
SSD~\cite{ssd} were designed for real-time computer vision applications that
require high accuracy in the presence of latency constraints. However, these
solutions are evaluated offline, one image at a time, and do not consider the
impact of increase in inference latency on the application performance. In
real-time systems such as autonomous vehicles, the models are deployed in an
online streaming setting where the ground truth changes during inference time as
shown in \Cref{f:tsp-figure}. To evaluate performance in streaming settings, Li
et al.~\cite{streaming-perception} proposed a modified metric that measures the
model performance against the ground truth at the end of inference. They evaluated
object detection models in a streaming fashion, and found that the
average precision of the best performing model drops from 38.0 to 6.2, and picking
the model that maximizes streaming average precision reduces the drop to 17.8.


%
We confirm the findings of Li et al.~\cite{streaming-perception}, and extend
their analysis to object tracking. The standard metric for object
tracking is \mota{} (multiple object tracking
accuracy)~\cite{milan2016mot16,leal2017tracking}, and we refer to its streaming
counterpart as \tmota{}. ~\Cref{f:t-mota-sync-mota} shows that larger models
with higher \mota{} deteriorate in \tmota{} for the Waymo
dataset~\cite{waymo-data} as the higher latency widens the gap
between ground truth between inference start and finish.
The \mota{} of the largest model (EfficientDet-D7x) is 38.9
while the \tmota{} is 16.2. The model that maximizes
\tmota{} is EfficientDet-D4 with \mota{} of 32.1 and \tmota{} of 26.7.

\begin{figure*}[tb]
  \begin{minipage}[c]{0.5\textwidth}
    \includegraphics[width=\textwidth]{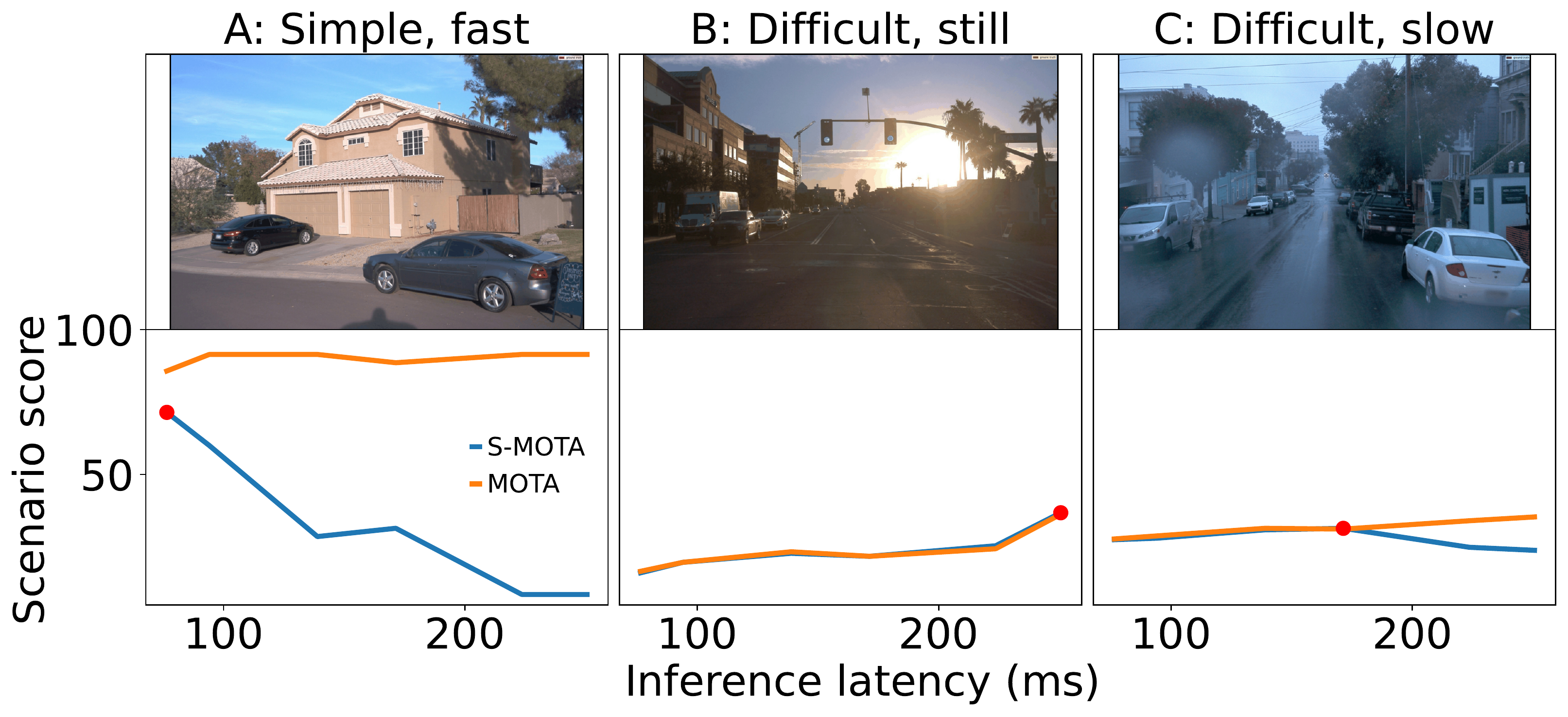}
  \end{minipage}\hfill
  \begin{minipage}[c]{0.45\textwidth}
    \caption{\small \textbf{Offline and streaming accuracies in three different
        scenarios.} While offline \mota{} (orange, y-axis) of the
      EfficientDet~\cite{tan20efficientdet} models increases as inference latency
      (x-axis) increases, streaming \mota{} (\tmota{} in blue, y-axis), responds
      depending on the scenario context (frames). Compare these plots
      to~\cref{f:t-mota-sync-mota}, which shows the same on average over the
      entire dataset.}
    \label{f:example-cases}
  \end{minipage}
\end{figure*}

Since environment context varies, to further analyze the tradeoffs between
latency and accuracy, we identify the model that maximizes the \tmota{} of
1-second video segment scenarios in the Waymo dataset. We observe that the best
performing model varies widely from scenario to scenario (\Cref{f:example-cases}).
Scenarios that are difficult (e.g., sun glare, drops
on camera and reflection) and still (e.g., standing cars in intersection) show
(\Cref{f:example-cases} center) benefit from stronger perception while incurring
marginal penalty from the latency increase. On the other hand, simple and fast
scenes with rapid movement (e.g., turning, in \Cref{f:example-cases} left),
behave the opposite, where performance degrades sharply with latency. As a
result, at the video segment-level the optimization landscape of streaming
accuracy looks vastly different than on aggregate
(\Cref{f:t-mota-sync-mota}).

In this paper, we propose leveraging contextual cues to optimize \tmota{}
dynamically at test time. The object detection model is just one of several
choices that we refer to as \textit{\params{}} to consider in an object
tracking system. Our method, called Octopus, optimizes \tmota{} at test time by
dynamically tuning the \params{}. Concretely, we train a light-weight
second-order model to switch between the \params{} using a battery of
environment features extracted from video segments, like obstacle movement
speed, obstacle proximity, and time of day.

Our contributions in this work can be summarized as:
\begin{tightitemize}
  \item We are the first to analyze object tracking in a streaming setting. We
  show that the models that maximize \tmota{} change per scenario and propose
  that the optimization tradeoffs are a result of scene difficulty as well as
  obstacle displacement.
  \item We present a novel method of \tmota{} optimization that leverages
  contextual features to switch the object tracking configuration at test-time.
  \item Our policy improves tracking performance (\tmota{}) by $7.4\%$ over a
  static approach by evaluating it on the Waymo dataset~\cite{waymo-data}. We
  improve \tmota{} by $3.4\%$ when we apply this approach on the Argoverse
  dataset~\cite{argo-data}.
\end{tightitemize}


%% file: related.tex
\section{Related Works}
\myparagraph{Latency \emph{vs.} accuracy} Prior works have recognized the
tension between latency and accuracy in perception
models~\cite{yolov4,pylot-paper,speed-accuracy-detectors,tan20efficientdet}, and
examined facets of the tradeoff between decision speed \emph{vs.}
accuracy~\cite{pleskac2010two}.
However, these works study this tradeoff in offline static settings. Li
et al~\cite{streaming-perception} examine this tradeoff in streaming settings,
where the ground truth of the world changes continuously. They show that
conventional accuracy misrepresents perception performance in such settings,
and proposed streaming accuracy. Their results were shown in
detection (as well as semantic segmentation~\cite{courdier2020real}), so we
first verified that the idea also holds in tracking. Further, while Li et al.
maximize streaming accuracy on average, we expose how the environment context
plays a crucial role on its behavior. We leverage the context
to dynamically maximize streaming accuracy at test time.
%

\myparagraph{Model serving optimization and dynamic test-time adaptation}
To reduce the inference latency, techniques such as model
pruning~\cite{han2015learning,lin2017runtime,luo2017thinet} and
quantization~\cite{park2017weighted,xu2018deep,zhao2019improving} have been
proposed.
While these techniques focus on effectively reducing the size of the models, our
focus is on the policy: when and where to do it, in order to optimize accuracy
in streaming settings. These techniques could be applied to reduce the latency
of the models we use.
%
Most similar to our setting are works that leverage context to dynamically
adjust model architecture at test time to reduce resource
consumption~\cite{jiang2018chameleon,wang2018skipnet} or to improve
throughput~\cite{shen2017fast}.
This work employs a similar approach in leveraging insight about latency
\emph{vs.} accuracy tradeoff in \av{} (autonomous vehicle) perception.



\myparagraph{Inferring configuration performance without running}
Several existing works achieve resource 
savings by modeling from data how a candidate
configuration would perform without actually running it.
Hyperstar~\cite{mittal2020hyperstar} learns to approximate how
a candidate hyperparameter configuration would perform on a dataset without
 training.
Chameleon~\cite{jiang2018chameleon} reduces profiling cost of 
configurations for surveillance camera detection by assuming temporal locality
among profiles.
One key differentiating factor with our work is that
the prior works were designed for a hybrid setting where some profiling is allowed.
In our case, the strict time and compute constraints in the \av{}
setting~\cite{driving-michigan} restrict this approach, requiring an
approximation-only method like \system{}.

%% file: problem-setup.tex
\section{Problem Setup}
\label{s:problem-setup}
We first lay out the problem formulation (\cref{ss:problem-formulation}).
Next, after an introduction of the \system{} dataset
(\cref{ss:octopus-dataset}), we measure the accuracy opportunity gap between the
global best policy~\cite{streaming-perception} and the optimal dynamic policy
(\cref{ss:opportunity-gap}).
%
%
Finally, we perform a breakdown analysis of the components needed in order to
optimize streaming accuracy at test time (\cref{ss:timely-accuracy-analysis}).

%
%


\subsection{Problem formulation}
\label{ss:problem-formulation}
Given real-time video stream as a series of images, we consider the inference
of an \av{} pipeline (obstacle detection and tracking) on this stream.
%
%
Let $\mathbb{S}$ denote the mean \tmota{} score of the tracking model, which
depends on the values of the \params{} $\mathcal{H}$, such as object
detection model architecture and maximum age of tracked objects.
Currently, the \params{} $h \in \mathcal{H}$ are chosen using offline datasets,
and are kept constant during deployment~\cite{streaming-perception}.
We refer to this method as the \emph{global best} approach for statically
choosing global \params{} ($h_{global}$), which are expected to be best across
all driving scenarios.

In contrast, in this work we study whether the \tmota{} score $\mathbb{S}$ can be improved by
dynamically changing $h$ every $\Delta \tau$ at test time.
The \params{} we choose at each time period
$[\tau, \tau+\Delta \tau)$ is $h_\tau$, and the corresponding score-optimal
values is $h^*_\tau$.

\subsection{The \system{} dataset}
\label{ss:octopus-dataset}
To generate the \system{} dataset ($\mathcal{D}$), we divide each video
of a driving dataset into consecutive segments of duration $\Delta \tau$.
We run the perception pipeline with a range of values of \params{}
$\mathcal{H}$, and record the \tmota{} score for each segment
$s^h_\tau$.
We assign the optimal $h^*_\tau$ to the \params{} that achieve
the highest \tmota{} score (i.e., the optimal \tmota{} score
$s^*_\tau$).
The segment duration $\Delta \tau$ is chosen to be short. This allows more
accurately studying the performance potential of dynamic streaming accuracy
optimization, because decision-making over smaller intervals generally performs
better.


%
We generate the \system{} dataset by recording \param{} values $h_\tau$ and the
corresponding \tmota{} scores $s^h_\tau$ of the Pylot \av{}
pipeline~\cite{pylot-paper} for the Argoverse and Waymo
datasets~\cite{argo-data,waymo-data}.
We execute Pylot's perception consisting of a suite of 2D object detection
models from the EfficientDet model family~\cite{tan20efficientdet} followed by
the Simple, Online, and Real-Time tracker~\cite{tracking_sort}.
For each video scenario, we explore the following \params{}:
\begin{tightitemize}
  \item{} \textbf{Detection model architecture}: selects the model from the
  EfficientDet family of models, which offers different latency \emph{vs.}
  accuracy tradeoff points.
  \item{} \textbf{Tracked obstacles' maximum age}: limits the duration for which
  the tracker continues modeling the motion of previously-detected obstacles,
  under the assumption of temporary occlusion or low detection confidence
  (flickering).

  %
\end{tightitemize}
Other \params{} had limited effect
on performance (\cref{app:other-metaparameters}).

We run every \param{} configuration in a Cartesian product of selected values
for each \param{}, and record latency metrics, detected objects, and tracked
objects.
The resulting $18$ \param{} configurations yield $\approx18,000$ trials. We 
make this dataset public (\url{https://github.com/EyalSel/Contextual-Streaming-Perception}).



\begin{table}[t]
  \scriptsize
  \centering
  \caption{\small \textbf{Dynamically changing \params{} creates an accuracy opportunity gap.}
    The \emph{streaming} accuracy (\tmota{}) of the \emph{global best} \params{}
    $h_{global}$ is $6.1$ points lower on average than that of the
    \emph{optimal} \params{} $h^*_\tau$. Similarly, there is a $3.2$ gap in
    \mota{} (top).}
  \input{tables/LAT_AR-optimization_gap.tex}
  \label{t:oracle-gap}
\end{table}

\subsection{Accuracy opportunity gap}
\label{ss:opportunity-gap}
%
In order to study if the \emph{global best} \params{} $h_{global}$ offer
the best accuracy in all driving scenarios, we split the \system{}
dataset ($\mathcal{D}$) into train ($\mathcal{D}_{train}$) and test
($\mathcal{D}_{test}$) sets.
%
Next, we compute global best \params{} ($h_{global}$) as the configuration that
yields the highest mean \tmota{} score across all video segments in
$\mathcal{D}_{train}$. We denote $h_{global}$'s mean \tmota{} scores on the
train and test set as $s^{global}_{train}$ and $s^{global}_{test}$,
respectively.
Similarly, we denote the mean \tmota{} scores of $h^*_\tau$ (i.e., optimally
changing \params{}) as $s^*_{train}$ and $s^*_{test}$.

\ionel{We don't explain at all why we've included \mota{} as well.}
We define the \emph{\tmota{} opportunity gap} between the optimal dynamic
\params{} and the global best \params{} as the upper bound of
$s^*_{test} - s^{global}_{test}$. We repeat the same calculation for \mota{}.
In \Cref{t:oracle-gap}, we show the opportunity gap for the Argoverse and Waymo
datasets~\cite{argo-data,waymo-data} using $\Delta \tau=1s$. We conclude that
optimally choosing the \params{} at test time offers a $6.1$ (Waymo) and $8.5$
(Argoverse) \tmota{} improvement on average, along with reductions in streaming
false positives/negatives and streaming ID switches.

Of note, if offline accuracy were to increase uniformly across all
configurations and scenarios, the performance improvement of the dynamic
approach over the static baseline is expected to persist.
This applies to the opportunity gap shown above (the optimal improvement), as
well as for any dynamic policy improvement in this space.
This means that performance improvement achieved by dynamic optimization come in
addition to, and not instead of, further advances in conventional (offline)
tracking.
%
%
\eyal{Writing note: Please keep in mind that there are other minor confounding
  factors beyond non-uniform increase that technically prevent making the last
  statement outright. The writing is meant to punt these issues to future work}

\subsection{Streaming accuracy analysis}
\label{ss:timely-accuracy-analysis}
We approach dynamic configuration optimization as a ranking
problem~\cite{liu2009learning}, and solve it by learning to predict the
difference in score of configuration pairs in a given scenario
context~\cite{yogatama2014efficient}.
We decompose this learning task into predicting the difference in \one{}
\mota{}, and \two{} accuracy degradation during inference.
%
%
To our knowledge we're the first to perform this analysis.

\textit{Decomposition.} Streaming accuracy (\tmota{}) is tracking accuracy
against ground truth at the end of inference, instead of the beginning
(\mota{})~\cite{streaming-perception}.
Offline accuracy (\mota{}) degrades as a result of change in ground truth during
inference.
The gap between \mota{} and \tmota{} is defined here as the ``degradation''.
Therefore, \tmota{} is expressed as $S-D$ where $S$ is \mota{} and $D$ is the
degradation.
The difference in \tmota{} between two configurations is $(S_1 - S_2) - (D_1 -
  D_2)$. This decomposition separates the difference in \tmota{} of two
configurations into two parts:
%
\begin{tightitemize}
\item $(S_1 - S_2)$ is the difference in offline accuracy. This
difference originates from the \mota{} boost, which is affected by \textit{(i)} the
added modeling capacity influenced by the scene difficulty for detection, a
specific case of the more general example difficulty~\cite{baldock2021deep},
and \textit{(ii)}  the max-age choice which depends on the scene obstacle
displacement~\cite{center_track}.
\item $(D_1 - D_2)$ is the difference in accuracy degradation.
This may be derived from: a. the difference configuration latencies and b.
scene obstacle displacement.
\end{tightitemize}

\textit{Predicting both components to optimize streaming accuracy.}
At test-time, $S_1 - S_2$ and $D_1 - D_2$ are predicted for each scenario.
We find that accurately predicting both components per environment context is
necessary to realize the \textit{opportunity gap} (see
\cref{ss:opportunity-gap}).
First, we show that perfectly predicting $D_1 - D_2$ ($\Delta D^*$)
and $S_1 - S_2$ ($\Delta S^*$) (\Cref{t:timely-acc:sync-deg}, top left) yields
$31.2$ \tmota{}, the same as the optimal policy $h_\tau^*$ on Waymo
(\Cref{t:oracle-gap} bottom panel, row 2).
Then, we predict $D_1 - D_2$ and $S_1 - S_2$ on average across all scenarios for
each configuration ($\overline{\Delta D}$ and $\overline{\Delta S}$ respectively
in~\Cref{t:timely-acc:sync-deg}). Combining $\overline{\Delta D}$ and
$\overline{\Delta S}$ (\Cref{t:timely-acc:sync-deg}, bottom-right) yields $25.1$
\tmota{}, the same as the global best policy $h_{global}$ score on Waymo
(\Cref{t:oracle-gap} bottom panel, row 1).
The hybrid-optimal policies (\Cref{t:timely-acc:sync-deg} top-right and
bottom-left) achieve $1.8$ ($26.9-25.1$) and $2.7$ ($27.8-25.1$) of the $6.1$
($31.2 - 25.1$) optimal policy opportunity gap.
Taken together, these results demonstrate the need to accurately predict both
the change in \mota{} ($S_1 - S_2$) and in degradation ($D_1 - D_2$) per
scenario in order to optimize \tmota{} at test time.

\begin{table*}[tb]
  \begin{minipage}[c]{0.2\textwidth}
    \begin{tabular}{@{}lcc@{}}
      \toprule
                            & $\Delta D^*$ & $\overline{\Delta
      D}$                                                      \\
      \midrule $\Delta S^*$ & 31.2         & 26.9              \\
      $\overline{\Delta S}$ & 27.8         & 25.1              \\
      \bottomrule
    \end{tabular}
  \end{minipage}\hfill
  \begin{minipage}[c]{0.77\textwidth}
    \caption{\small \textbf{Both streaming accuracy components must be predicted
        per scenario in order to realize the full dynamic policy's opportunity gap.}
      $\Delta D^*$ is optimal degradation prediction, $\Delta S^*$ is optimal
      offline gap prediction. $\overline{\Delta D}$ and $\overline{\Delta S}$ are
      the corresponding global-static policies.}
    \label{t:timely-acc:sync-deg}
  \end{minipage}
\end{table*}

%% file: tables/LAT_AR-optimization_gap.tex
\setlength\tabcolsep{2.5pt}
\begin{tabular}{@{}l|l|ccccc}
    \toprule
    \textbf{Method} & \textbf{Dataset} & \textbf{\mota{}}$\uparrow{}$  & \textbf{\motp{}}$\uparrow{}$  & \textbf{FP}$\downarrow{}$     & \textbf{FN}$\downarrow{}$     & \textbf{ID\textsubscript{sw}}$\downarrow{}$     \\ \midrule
    Global best     & Waymo            & 37.3                          & 78.1                          & 21515                         & 543193                        & 11615                                           \\
    Optimal         & Waymo            & 40.5                          & 77.6                          & 15738                         & 532678                        & 9137                                            \\
    Global best     & Argoverse        & 63.0                          & 82.1                          & 6721                          & 43376                         & 1284                                            \\
    Optimal         & Argoverse        & 70.8                          & 81.0                          & 5772                          & 34210                         & 828                                             \\ \bottomrule
    \toprule
    \textbf{Method} & \textbf{Dataset} & \textbf{\tmota{}}$\uparrow{}$ & \textbf{\tmotp{}}$\uparrow{}$ & \textbf{\tfp{}}$\downarrow{}$ & \textbf{\tfn{}}$\downarrow{}$ & \textbf{\tid{}\textsubscript{sw}}$\downarrow{}$ \\ \midrule
    Global best     & Waymo            & 25.1                          & 72.2                          & 33616                         & 633159                        & 11212                                           \\
    Optimal         & Waymo            & 31.2                          & 71.0                          & 28907                         & 590847                        & 6997                                            \\
    Global best     & Argoverse        & 49.4                          & 75.2                          & 13485                         & 55484                         & 1092                                            \\
    Optimal         & Argoverse        & 57.9                          & 74.1                          & 9562                          & 48354                         & 708                                             \\ \bottomrule
\end{tabular}


%% file: methods.tex
\section{\system{}: Environment-Driven Perception}
\label{s:policy}
%
We propose leveraging properties of the \av{} environment context (e.g., ego
speed, number of agents, time of day) that can be perceived from sensors in
order to dynamically change \params{} at test time.
We first formally present our approach for choosing \params{}, which uses
regression to infer a ranking of \param{} configurations
(\cref{ss:policy-regression}).
Then, we describe the environment representation that allows to effectively
infer each component of the streaming accuracy~(\cref{ss:policy-environment}).
%

\subsection{Configuration Ranking via Regression}
\label{ss:policy-regression}
In order to find the \params{} $h_\tau$ that maximize the \tmota{} score
$s^h_\tau$ for each video segment, \system{} first learns a regression model
$M$.
%
%
The model predicts $s^h_\tau$ given the \params{} $h_\tau$ and the
representation of the environment $e_\tau$ for the period $\tau$.
Following, \system{} considers all \param{} values, and picks the \params{} that
give the highest predicted \tmota{}.

%
%
Executing the model $M$ at the beginning of each segment $\tau$ requires an
up-to-date representation of the environment.
However, building a representation requires the output of the perception
pipeline (e.g., number of obstacles).
\system{} could run the current perception configuration in order to update the
environment before executing the model $M$, but this approach would greatly
increase the response time of the \av{}.
Instead, \system{} makes a Markovian assumption, and inputs the environment
representation of the previous segment $e_{\tau-1}$ to the model $M$.
\begin{equation}
  M(h_\tau, e_{\tau-1})  =  \hat{s}^h_\tau \label{eq:regression-task}
\end{equation}
Thus, it is important to limit the segment length $\Delta \tau$ as the longer a
segment is, the more challenging it is to accurately predict the scores due to
using an older environment representation (see~\cref{ss:eval-setup} for our
methodology for choosing $\Delta \tau$).

The model is trained using the mean squared error loss.
%
We choose the \params{} that the model predicts as the highest score.
%
Let $s^{global}_\tau$ denote the \tmota{} score obtained with $h^{global}$
\params{}, which by definition is a lower bound of the optimal \tmota{} score
$s^*_\tau$ (i.e., $s^{global}_\tau \leq s^*_\tau$).
Thus, in order to pick the best \params{}, \system{} can only predict $s^h_\tau$
relative to $s^{global}_\tau$.
As a result, \system{} utilizes the following as the final loss:
\begin{equation}
  \label{eq:final_loss}
  L  = \frac{1}{N} \sum_{i=0}^{N} \left( \hat{r}^h_i - clip((s^h_i - s^{global}_i), \epsilon) \right)^2
\end{equation}
\justin{is this $\hat{r}^h_i$ suppose to be the "clipped relative score"} where
N is the size of the training data and $\hat{r}^h_i$ is the relative \tmota{}
score predicted by the model $M$,
$
  \hat{r}^h_i = \hat{s}^h_i - s_i^{global}
$.

Finally, \system{} clips by lower bounding the predictions by $s^{global}_i -
  \epsilon$ as the predictions significantly below $s_i^{global}$ are irrelevant
to the optimization problem.
Moreover, by clipping the predictions, \system{} reduces the dynamic range of
the regressor and makes it easier to predict the higher \tmota{} scores.




\subsection{Environment Representation}
\label{ss:policy-environment}
We can represent the environment context $e_\tau$ by capturing the
characteristics of the video segment.
In order to keep the decision-making latency small, we eschew more complex
learned representation designs.
%
Instead, we developed hand-engineered features from sensors and the outputs of
the object detection model following prior work by Nishi et
al.~\cite{nishi2020fine}, where the features were used to predict human driving
behavior.
We collect the features per frame and then aggregate by averaging across frames
in the video segment.
Unless otherwise specified, we use the 10\thss{} percentile, mean, and 90\thss{}
percentile of the following features:
\begin{tightenumerate}
  \item \textbf{Bounding box speed} is the distance traveled by an object in
  pixel space across two frames. This feature infers the speed of objects, and
  thus it is important for capturing obstacle displacement to predict the
  streaming accuracy degradation.
  %
  \item \textbf{Bounding box self IoU} is the Intersection-over-Union (IoU) of
  an object's bounding box in the current frame relative to the previous frame.
  \ionel{Is it always the previous frame?}
  Along with the bounding speed, this feature helps isolate the change in size
  of an obstacle, as it moves towards or away from the ego vehicle. \ionel{How
    does this feature isolate the change in size? The IoU could also be small if
    the object moves quickly.}
  \item \textbf{Number of objects} is measured per frame, and indicates the
  complexity of a scene as the more objects in a scene, the more likely the
  \av{} is to encounter object path crossings and occlusions (scene
  difficulty). Therefore, this feature signals when to prioritize for high
  offline accuracy configurations that are robust to object occlusions.
  %
  \item \textbf{Obstacle longevity} is the number of frames for which an
  obstacle has been tracked. Lower obstacle longevity implies more occlusion as
  obstacles enter and leave the scene, making perception more difficult. This
  feature also guides the choice of tracking \params{} as low obstacle
  longevities correlate with lower tracking maximum age giving better
  performance.
  \item \textbf{Ego driving speed} is the speed at which the \av{} is traveling,
  and indicates the environment in which the \av{} is driving (e.g., highway vs.
  city).
  \item \textbf{Ego turning speed} is the angle change of the \av{}'s direction
  between consecutive frames. This feature helps differentiate the source of
  apparent obstacle displacement between obstacle movement and ego movement.
  %
  \item \textbf{Time of day} hints if a high-accuracy configuration is required
  to handle challenging scenes (e.g., night driving, sun glare at sunset).
  \ionel{Nit: might want to specify how it is measured.}
\end{tightenumerate}

We compute these feature statistics for different bounding box size ranges in
order to reason about obstacle behavior depending on the detection model
strength needed for their accurate detection.
For example, higher offline accuracy models do not confer better streaming
accuracy if the obstacles in the detectable size range move too rapidly for the
longer detection inference time to keep up with.


%
Note that the choice of \params{} in each video segment changes the
configuration, and hence the outputs of the tracking pipeline.
Therefore, the resulting environment representation is no longer independently
and identically distributed (i.i.d), making it challenging to use traditional
supervision techniques.
To keep the data distribution stationary during train time, we use the object
detection model as given by the global best \params{} $h_{global}$.
While this training objective is biased considering that the features may be
derived from any configuration, we find that it reduces variance during
training, and works better than using features derived from the ground
truth.

We concatenate the \params{} $h$ and the environment representation vector
$e_\tau$, and then use a supervised regression model to predict the score,
optimizing using the loss given in \cref{eq:final_loss}.

In addition, we compare our method to a conventional CNN model approach
(ignoring its resource and runtime requirements) in
\cref{app:neural-network-policy}.

%% file: evaluation.tex
\section{Experiments}
\label{s:evaluation}
Next, we analyze our proposed approach. The subsections discuss the following:
\begin{tightenumerate}
  \item \textbf{Setup (\cref{ss:eval-setup})}: Description of Datasets and Model
  Details
  \item \textbf{Main results (\cref{ss:eval-main-results})}: What is the
  performance improvement conferred by the dynamic policy over the static
  baseline?
  \item \textbf{Explainability (\cref{ss:eval-explainability})}: \textbf{I.}
  Does the learned policy behavior match human understanding of the driving
  scenario? \textbf{II.} Where is the learned policy similar to and different
  from the optimal policy? \textbf{III.} How are scenarios clustered by
  \param{} score? \textbf{IV.} What is the relative importance of the
  hand-picked features and of the \params{} towards \tmota{} optimization?
  \item \textbf{Ablation study (\cref{app:eval-ablations-B})}: How do various
  ranking implementation choices affect the final performance?
\end{tightenumerate}
\subsection{Methodology}
\label{ss:eval-setup}
We evaluate on the Argoverse and Waymo datasets~\cite{argo-data,waymo-data},
covering a variety of environments, traffic, and weather conditions.
We do not use the private Waymo test set as it does not
support streaming metrics.
However, we treat the Waymo validation set as the test set, and we
perform cross validation on the training dataset ($798$ videos for training,
and $202$ videos for validation).
Similarly, we divide the Argoverse videos into $75$ videos for training, and
$24$ for validation.
In addition, we follow the methodology from Li et.
al~\cite{streaming-perception} in order to create ground truth 2D bounding
boxes and tracking IDs, which are not present in the Argoverse dataset.
We generate these labels using QDTrack~\cite{qdtrack} trained on the BDD100k
dataset, which is the highest offline accuracy model available.
%

%
In our experiments, we run object detectors from the EfficientDet
architecture~\cite{tan20efficientdet} on NVIDIA V100 GPUs, and the SORT
tracker~\cite{tracking_sort} on a CPU (simulated evaluation on faster hardware
is in \cref{app:faster-hardware-simulation}). We chose to use the EfficientDet
models because they are especially optimized for trading off between latency
and offline accuracy, and because they are close to the state-of-the-art.
%
The EfficientDet models were further optimized using Tensor
RT~\cite{tensor-rt}, which both reduces inference latencies to less than
$250$ms (the latency of the largest model, EfficientDet-D7x) and decreases
resource requirements to at most $3$ GPUs.
We use these efficient models to investigate the effect of optimizing two
\params{}: detection model and tracking max age (see~\cref{ss:octopus-dataset}
for details).
We compile $18$ configurations of the \av{} perception pipeline by exploring
the Cartesian product of the values of the two \params{}
(see \Cref{a:config-params}).

\begin{lstlisting}[language=Python,
    keywordstyle=\color{black},
    commentstyle=\color{codegreen},
    caption={\small \textbf{Values for detection and tracking \params{}.}},
    label=a:config-params]
# Metaparameters
detection-model = {EfficientDet: 3, 4, 5, 6, 7, 7x}
tracking-maximum-age = {1, 3, 7}
\end{lstlisting}

We implement \system{}'s policy regression model as a Random
Forest~\cite{breiman2001random} using \cref{eq:final_loss} to choose \params{}
for video segment length $\Delta \tau=1s$.
%
%
We describe training details in \cref{app:training-details}.
%

\myparagraph{Policy Runtime Overhead}
Every step $\Delta \tau$, the \system{} policy applies Random Forest regression
for all the $18$ perception pipeline configurations in order to predict the
best one to apply in the next step.
Due to the lightweight policy design, inference on all $18$ configurations has
a latency of at most $6$ms using a single Intel Xeon Platinum 8000 core.
As a result, the policy decision finishes before the sensor data of the next
segment arrives, and thus does not affect latency of the perception pipeline.
Moreover, \system{} pre-loads the perception model weights (13.86GB in total in
our experiments) and forward pass activations in the GPU memory ($32$GB), and
thus actuates \param{} changes quickly.


\input{evaluation-main-results}

\input{evaluation-explainability}




%% file: evaluation-main-results.tex
\subsection{Main results}
\label{ss:eval-main-results}
We compare the \system{} policy with the optimal policy ($h^*_\tau$) and the
global best policy ($h^{global}$) as proposed by Li et
al.~\cite{streaming-perception}.
In addition, following the discussion in \cref{ss:policy-environment}, we show
ablations of the \system{} policy in order to highlight the relative
contribution in both the setting of predictive and close-loop \param{}
optimization.
%
Concretely, we include the following setups that \system{} can use to optimize
the \params{} at time $t$:
\begin{tightitemize}
  \item{} \emph{Ground truth from current segment}: features are derived from
  the sensor data and the labels at time $t$.
  \item{} \emph{Ground truth from previous segment}: features are derived from the
  sensor data and the labels at time $t - \Delta \tau$.
  \item{} \emph{Closed-loop prediction from previous segment}: features are derived
  from the sensor data and the output of the perception pipeline for the
  previous segment (i.e., at time $t - \Delta \tau$).
\end{tightitemize}

In \Cref{t:main-results} we show the tracking accuracy results for the Waymo and
Argoverse datasets.
The results show that the \system{} policy with closed-loop prediction
outperforms the global best policy by $1.9$ \tmota{} (Waymo) and
$1.7$ \tmota{} (Argoverse).
Both the optimal and \system{} policies achieve further accuracy increases using
features derived from the current segment, further illustrating how rapidly
configuration score changes over time.
The consistent accuracy improvements across the two datasets show
that leveraging environment context to dynamically optimize streaming
accuracy provides substantial improvements over the state-of-the-art.
Moreover, as discussed in~\cref{ss:opportunity-gap}, streaming accuracy
improvement over the global best policy will likely persist independently of
innovation in offline accuracy of the underlying perception models.


%

\begin{table}[tbh]
  \scriptsize
  \centering
  \caption{\small \textbf{Streaming tracking accuracy results on two datasets.}}
  \begin{subtable}{1\columnwidth}
    \centering
    \caption{\small Waymo}
    \input{tables/LAT_AR-waymo-full.tex}

  \end{subtable}
  \begin{subtable}{1\columnwidth}
    \centering
    \caption{\small Argoverse}
    \input{tables/LAT_AR-argo-full.tex}

  \end{subtable}
  \label{t:main-results}
\end{table}

\myparagraph{\tmota{} \emph{vs.} \tmotp{}}
\justin{this is way to long; we should instead get to our point.
1. \tmota{} is the metric we chose to optimize.
2. \tmotp{} is naturally goes down when \tmota{} goes up
(technically not true, cause you can have a better model due to better training data/augmentation).
3. You could optimize \tmotp{} if you wanted, look at the supplement.}
\Cref{t:main-results} highlights that both \system{} and optimal policy
occasionally deteriorate \tmotp{}, inversely to the improvement in
\tmota{}.
This result reflects on a broader pattern where in streaming settings bounding
boxes in general lag after the ground truth, even if by a small enough margin
to be counted as true positives. \tmotp{}, which is weighted by the IOU between
correct predictions and the ground truth is especially hurt as a result.
\tmota{}, which just counts the number of false positives, is less affected by
this. We provide a longer analysis of this tradeoff/pareto-frontier between
\tmota{} and \tmotp{} in \cref{app:mota-motp}.

%

%% file: tables/LAT_AR-waymo-full.tex
{
\begin{tabular}{@{}l|ccccc@{}}
    \toprule
    \textbf{Method}                        & \textbf{\tmota{}}$\uparrow{}$ & \textbf{\tmotp{}}$\uparrow{}$ & \textbf{\tfp{}}$\downarrow{}$ & \textbf{\tfn{}}$\downarrow{}$ & \textbf{\tid{}\textsubscript{sw}}$\downarrow{}$ \\ \midrule
    Global best                            & 25.1                          & 72.2                          & 33616                         & 633159                        & 11212                                           \\ \midrule
    Optimal                                & 31.2                          & 71.0                          & 28907                         & 590847                        & 6997                                            \\
    Optimal from the prev. segment         & 27.2                          & 71.2                          & 38631                         & 603777                        & 8092                                            \\ \midrule
    \textbf{\system{}} with:               &                               &                               &                               &                               &                                                 \\
    Ground truth from current segment      & 27.9                          & 72.3                          & 31489                         & 608870                        & 8966                                            \\
    Ground truth from prev. segment        & 27.3                          & 72.7                          & 31056                         & 612862                        & 9103                                            \\
    \textbf{Prediction from prev. segment} & \textbf{27.0}                 & \textbf{72.8}                 & \textbf{30272}                & \textbf{615780}               & \textbf{9511}                                   \\ \bottomrule
\end{tabular}
}

%% file: tables/LAT_AR-argo-full.tex
{
\begin{tabular}{@{}l|ccccc@{}}
    \toprule
    \textbf{Method}                         & \textbf{\tmota{}}$\uparrow{}$ & \textbf{\tmotp{}}$\uparrow{}$ & \textbf{\tfp{}}$\downarrow{}$ & \textbf{\tfn{}}$\downarrow{}$ & \textbf{\tid{}\textsubscript{sw}}$\downarrow{}$ \\ \midrule
    Global best                             & 49.4                          & 75.2                          & 13485                         & 55484                         & 1092                                            \\ \midrule
    Optimal                                 & 57.9                          & 74.1                          & 9562                          & 48354                         & 708                                             \\
    Optimal from the prev. segment          & 51.9                          & 74.0                          & 12652                         & 52804                         & 810                                             \\ \midrule
    \textbf{\system{}} with:                &                               &                               &                               &                               &                                                 \\
    Ground truth from current segment       & 53.2                          & 74.9                          & 11447                         & 52638                         & 917                                             \\
    Ground truth from prev. segment         & 51.6                          & 75.1                          & 11348                         & 54502                         & 1010                                            \\
    \textbf{Prediction from prev. segment } & \textbf{51.1}                 & \textbf{74.9}                 & \textbf{12062}                & \textbf{54469}                & \textbf{1008}                                   \\ \bottomrule
\end{tabular}
}

%% file: evaluation-explainability.tex
\subsection{Explainability}
\label{ss:eval-explainability}
We survey various aspects of the \param{} optimization problem, and
qualitatively compare the global best (baseline), the optimal, and the
\system{} (learned) policies.

\begin{figure*}[tb]
  \centering
  \includegraphics[width=0.99\textwidth, trim={4cm 18cm 8cm
        0},clip]{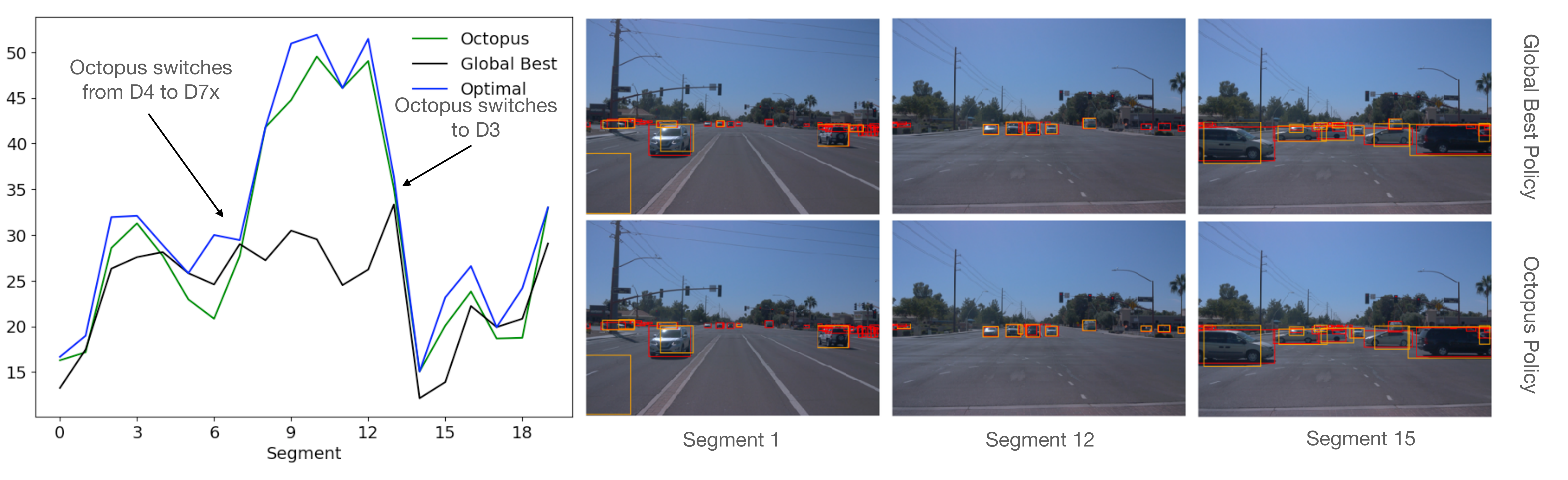} \caption{\small
        \textbf{Busy Intersection Scenario.} Left: The \tmota{} score of each
        $\tau=1s$ segment for the \system{} (learned), global best, and optimal
        policies. Right: The front-facing camera feed. Red bounding-boxes
        represent the ground truth, and orange represent the pipeline's
        predictions using the policy's configuration choice.}
  \label{fig:case-study}
\end{figure*}

\myparagraph{I. Case study: Busy Intersection}
\Cref{fig:case-study} shows a scenario where the ego vehicle enters a busy
intersection. The scenario is divided into three phases, where the learned (\system{})
policy adaptively tunes the \param{} to varying road conditions similarly to the
optimal policy plan.
First, the ego vehicle approaches the intersection with oncoming traffic on the
left. The vehicles in the distance cannot be picked up by any of the candidate
models. The learned policy chooses D4, the same model as the global best
policy.
Then, as the light turns yellow and the ego vehicle comes to a stop first in
the queue, the learned policy adapts by increasing the strength of
the model to D7x. The perception pipeline is now able to detect the smaller
vehicles in the opposing lane and adjacent to the road. The optimal policy
makes a similar decision while the global best policy remains with the D4
choice, incurring a net performance loss.
Finally, as vehicles in the cross-traffic start passing and occluding the
vehicles in the background, the learned policy returns to the lower latency D3
model.
This scenario illustrates how the learned policy can select the best model 
in each phase of the scenario, adapting to the change in driving environment.

\begin{figure*}[tb]
  \begin{minipage}[c]{0.5\textwidth}
  \includegraphics[width=\textwidth]{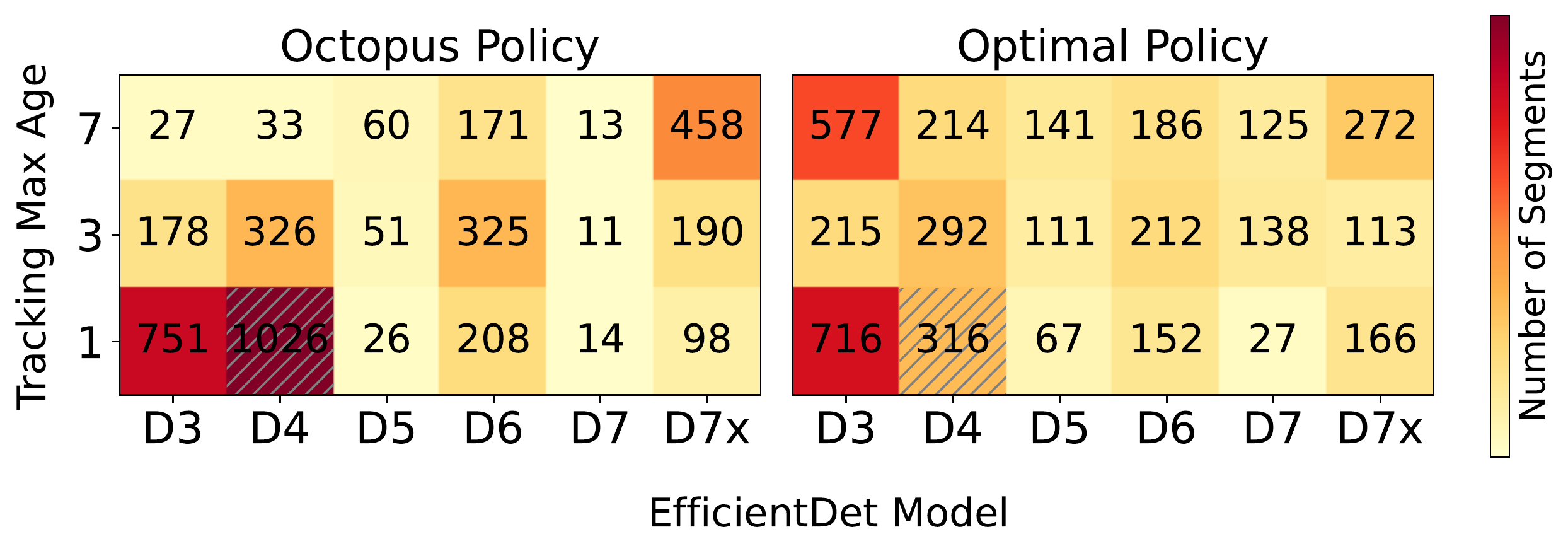}
  \end{minipage}\hfill
  \begin{minipage}[c]{0.47\textwidth}
    \caption{\small \textbf{Policy decision frequency}. The configuration choice
    frequency (color intensity) of the \system{} policy (left) and of the
    optimal policy (right). The global best configuration is
    EfficientDet-D4 with tracking max age of 1, emphasized with gray stripes.}
  \label{f:heatmap-policies}
  \end{minipage}
\end{figure*}

\myparagraph{II. Policy action heatmap}
In~\Cref{f:heatmap-policies}, we visualize the learned policy in comparison to
the optimal policy.
As we expect, the learned policy often selects the global best configuration
(D4-1), but expands out similarly to the optimal policy when performance can be
improved by changing configurations.
This result illustrates how, contrary to common belief that onboard perception
must have low latency (e.g. under 100ms \cite{waymo-data}), higher perception
latencies are tolerable and even preferable in certain environments. The
\system{} policy learns to take advantage of this when it opts for higher
offline accuracy models.
\eyal{TODO how do we explain the deviations from the optimal policy? Ablating
delta S with delta d hat on the t-SNE could reveal some aspects}
\justin{Honestly, not necessary; put this in the appendix if you must.}


\myparagraph{III. Scenarios are clustered by \param{} score}

Here, we evaluate whether scenarios are grouped by common \param{} score
behaviors. To this end, we first visualize the scenario score space (explained
below) as a t-SNE plot and then perform more formal centroid analysis.

To this end, each video segment of length $\Delta \tau=1s$ is vectorized by
computing the \mota{} score difference ($S_c - S_{h_{global}}$) and degradation
difference ($D_c - D_{h_{global}}$) for every \param{} configuration $c$ and the
global best configuration $h_{global}$ (see~\cref{ss:timely-accuracy-analysis}).
These values are then concatenated and normalized (z-score) across each
scenario.

\begin{figure*}[tb]
  \begin{minipage}[c]{0.57\textwidth}
  \includegraphics[width=\textwidth]{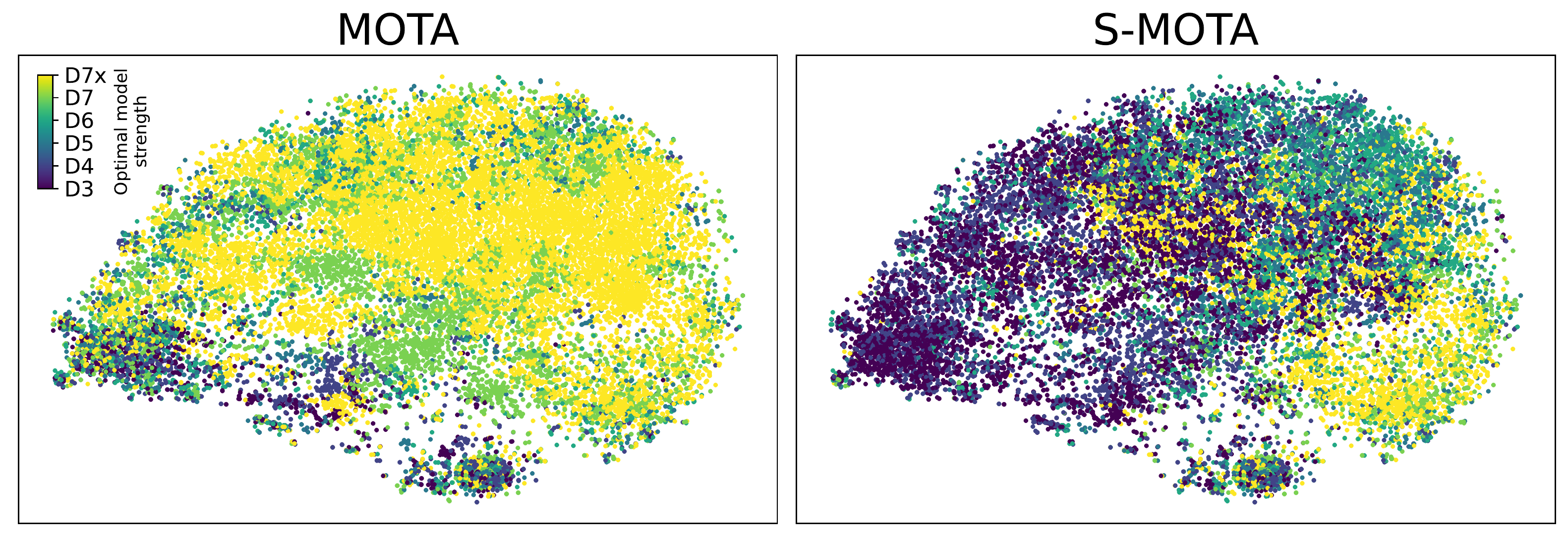}
  \end{minipage}\hfill
  \begin{minipage}[c]{0.4\textwidth}
    \caption{\small \textbf{The impact of the accuracy degradation incurred in
    online streaming context is scenario dependent.} Each point is a 1s driving
    scenario. Its color is the model that maximizes offline accuracy (left) and
    streaming accuracy (right).}
    \label{f:tsne-sync-timely}
  \end{minipage}
\end{figure*}

\textit{Cluster visualization.}
We visualize this space in a t-SNE plot in~\Cref{f:tsne-sync-timely}. The
scenario segment points are colored according to the model that optimizes
\mota{} (left) and \tmota{} (right).
We observe a non-uniform impact of accuracy degradation on the optimal model
choice in different video segment regions. This varying effect illustrates that
\tmota{} deviates from \mota{} setting on an environment context-dependent
basis. For case-study analysis of points in the t-SNE please
see~\cref{app:t-sne-visualization}.

\textit{Centroid analysis.}
We now perform a more formal clustering analysis that reveals modes of \param{}
optimality.
%
%

\begin{figure*}[tb]
  \begin{minipage}[c]{0.5\textwidth}
  \includegraphics[width=\textwidth]{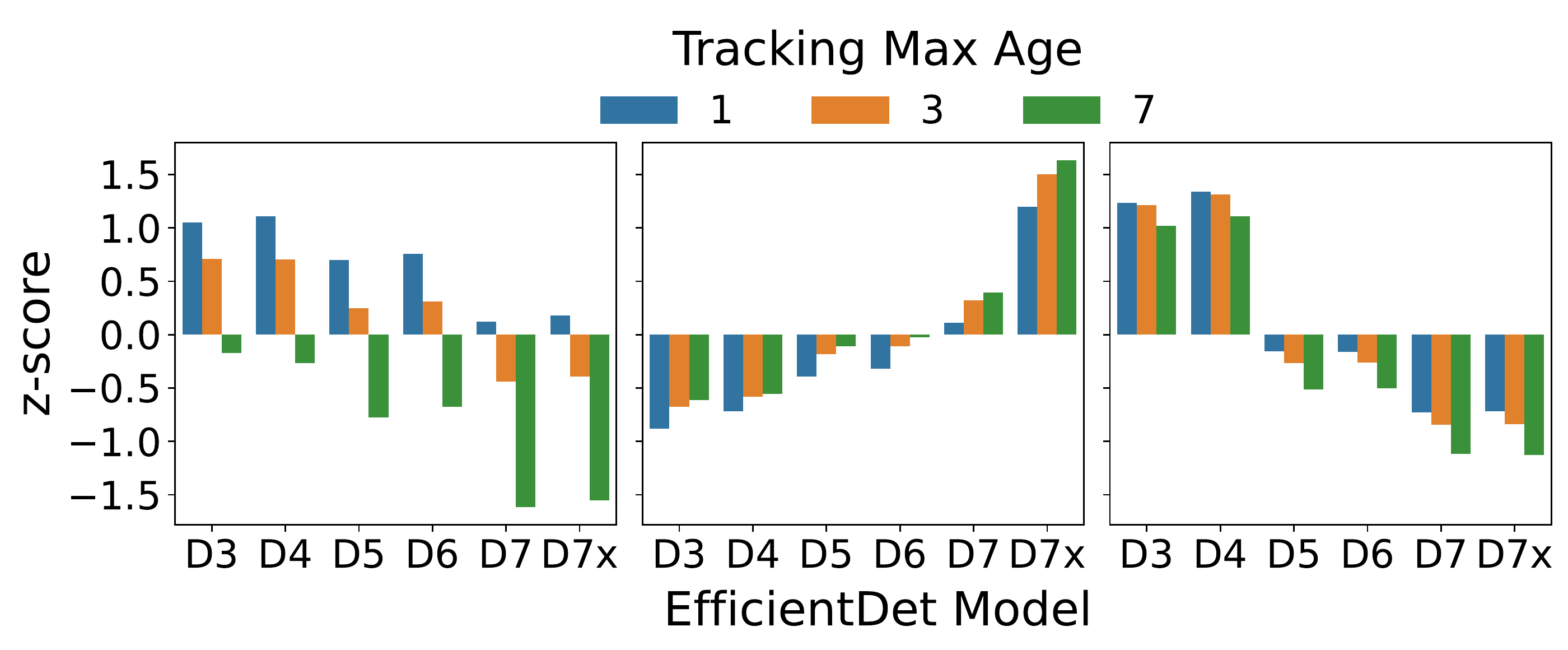}
  \end{minipage}\hfill
  \begin{minipage}[c]{0.47\textwidth}
    \caption{\small \textbf{Three centroids of the configuration score
    distribution}: \one{} preference for low max age (left), \two{}
    preference for offline accuracy (middle), and \three{} preference for low latency
    (right). The three clusters account for 34\% of the dataset.}
    \label{f:score-space}
  \end{minipage}
\end{figure*}

%
We perform K-means clustering, with k=8 on the z-score space.
\Cref{f:score-space} shows centroids of three representative scenario
clusters (right to left):
\one{} scenarios that benefit from lower latency detection.
\two{} scenarios that benefit from higher accuracy detection, and
\three{} scenarios that primarily benefit from a lower tracking max age, and to
a lesser extent from lower latency detection.
These distinct modes reflect that scenarios are grouped around similar \param{}
behaviors, corroborating with variation in scene difficulty and
obstacle displacement.
The rest of the scenario cluster visualizations are
in~\cref{app:centroid-visualization}.
\justin{State the conclusion
you want to draw; remove half the details. I assume you want to say:
"Intuitively, scenarios with similar preferences for models and max ages are
clustered together..." Not sure why I care about this as a reader.}


\begin{table}[]
  \scriptsize
  \centering
  \caption{\small \textbf{Importance study}}
  \begin{subtable}{1\columnwidth}
    \centering
    \caption{\small Feature importance (Regression)}
    \input{tables/feature-importance}
    \label{t:feature-importance}
  \end{subtable}
  \begin{subtable}{1\columnwidth}
    \centering
    \caption{\small \Param{} importance (Regression)}
    \input{tables/LAT_AR-metaparameter-importance.tex}
    \label{t:metaparameter-ablation}
  \end{subtable}
\end{table}

\myparagraph{IV. Feature and configuration \param{} importance}

\textit{Feature importance.}
To study the relative importance of the environment features used in our solution
(described in \cref{ss:policy-environment}), we show the feature importance
scores derived from the trained Random Forest regressor in
\autoref{t:feature-importance}.
The mean bounding box self-IOU and speed together constitute over half of the
normalized importance score, as they are predictive of the accuracy degradation
that higher-latency detection models would incur.
Aggregate bounding box statistics that capture scene difficulty, such as the
average number and longevity of the bounding boxes per frame, are also useful
for prediction.

\textit{Configuration \param{} importance.} In
\autoref{t:metaparameter-ablation}, we evaluate the gains attributed to each
\param{} by only optimizing one \param{} at a time, fixing the other parameter
to the value in the global best configuration. We do this to ablate the
performance shown in \Cref{t:main-results}, where they are optimized together.
In both cases, the models were trained on the ground truth and the present.
%
%
Although the detection model choice is the primary contributor, 
the additional choice of occlusion tolerance (max age)
 further contributes to the achieved performance.
We also observe how the performance gain when optimizing the tracker's maximum
occlusion tolerance (see~\cref{s:problem-setup} for details) does not combine
additively with the improvement in the detection model optimization.


%% file: tables/feature-importance.tex
\newcolumntype{?}{!{\vrule width 1pt}}
{
\setlength{\tabcolsep}{1.25pt}
\begin{tabular}{l|c?c|c}
    \toprule
    \textbf{Feature}          & \textbf{Importance Score} & \textbf{Feature}          & \textbf{Importance Score} \\ \midrule
    Mean BBox self IOU        & 0.304                     & BBox bin [665, 1024)      & 0.087                     \\
    Mean BBox Speed           & 0.200                     & BBox bin [1024, 1480)     & 0.071                     \\
    Num. BBoxes               & 0.063                     & BBox bin [1480, 2000)     & 0.055                     \\
    BBox Longevity            & 0.076                     & BBox bin [2000, 2565)     & 0.040                     \\
    Ego movement              & 0.096                     & BBox bin [2565, $\infty$) & 0.417                     \\
    Time of Day               & 0.008                     &                           &                           \\ \midrule
    \bottomrule
\end{tabular}
}

%% file: tables/LAT_AR-metaparameter-importance.tex
{
\setlength\tabcolsep{1.25pt}
\begin{tabular}{l|ccccc}
   \toprule
   \textbf{\Param{}} & \textbf{\tmota{}}$\uparrow{}$ & \textbf{\tmotp{}}$\uparrow{}$ & \textbf{\tfp{}}$\downarrow{}$ & \textbf{\tfn{}}$\downarrow{}$ & \textbf{\tid{}\textsubscript{sw}}$\downarrow{}$ \\ \midrule
   Global best       & 25.1                          & 72.2                          & 33616                         & 633159                        & 11212
   \\
   Detection-model   & 27.6                          & 72.4                          & 26116                         & 615771                        & 11254                                           \\
   Tracking-max-age  & 25.6                          & 71.9                          & 39400                         & 625184                        & 8973                                            \\
   Both              & 27.9                          & 72.3                          & 31489                         & 608870                        & 8966                                            \\
   \bottomrule
\end{tabular}

}

%% file: conclusions.tex
\section{Conclusions}
\label{s:conclusions}

Streaming accuracy is a much more accurate representation of tracking for
real-time vision systems because it uses ground truth at the end of inference
to measure performance. In this study we show the varying impact that
environment context has on the deviation of streaming accuracy from offline
accuracy. We propose a new method, \system{}, to leverage environment context
to maximize streaming accuracy at test time. Further, we decompose streaming
accuracy into two components: difference in offline accuracy \mota{} and the
degradation, and show that both must be inferred in every scenario to achieve
optimal performance. \system{} improves streaming accuracy over the global best
policy in multiple autonomous vehicle datasets.

\myparagraph{Acknowledgements} We thank Daniel Rothchild and Horia Mania for
helpful discussions.

%% file: appendix.tex
\section*{Supplementary Material: Table of contents}
\begin{itemize}
  \item \Param{} choice (\cref{app:other-metaparameters})
  \item Training details (\cref{app:training-details})
  \item Training hyperparameters (\cref{app:training_hparams})
  \item \tmota{} vs \tmotp{} (\cref{app:mota-motp})
  \item Study-case video (\cref{app:study-case-video})
  \item t-SNE visualization (\cref{app:t-sne-visualization})
  \item Full Centroid Visualization (\cref{app:centroid-visualization})
  \item Ranking Implementation Ablations (\cref{app:eval-ablations-B})
  \item Policy Design Using Neural Networks (\cref{app:neural-network-policy})
  \item Faster hardware simulation (\cref{app:faster-hardware-simulation})
\end{itemize}

\section{\Param{} choice}
\label{app:other-metaparameters}
\input{appendix-other-metaparams.tex}

\section{Training details}
\label{app:training-details}
\input{appendix-training-details.tex}

\section{Training hyperparameters}
\label{app:training_hparams}
\input{appendix-hyperparameters.tex}

\section{\tmota{} vs \tmotp{}}
\label{app:mota-motp}
\input{appendix-mota-motp.tex}

\section{Study-case video}
\label{app:study-case-video}
\input{appendix-study-case.tex}

\section{t-SNE visualization}
\label{app:t-sne-visualization}
\input{appendix-tsne-visualization.tex}

\section{Full Centroid Visualization}
\label{app:centroid-visualization}
\input{appendix-centroid-viz.tex}

\section{Ranking Implementation Ablations}
\label{app:eval-ablations-B}
\input{evaluation-ablations-B.tex}

\section{Policy Design Using Neural Networks}
\label{app:neural-network-policy}
\input{appendix-policy-neural-network.tex}

\section{Faster hardware simulation}
\label{app:faster-hardware-simulation}
\input{appendix-faster-hardware-simulation.tex}


%% file: appendix-other-metaparams.tex
In this study we use two \params{} (detection model and tracking maximum age)
out of several that we considered initially. These two \params{} were chosen
because \one{} they represent 83\% of the opportunity gap conferred by all the
\params{} considered together (see below) and \two{} a limited number of
\params{} allows computing the \system{} dataset on more scenarios.

\noindent\textbf{I. \Params{} we did not use:}
\begin{tightitemize}
    \item \textbf{Detection confidence threshold}: sets the minimum confidence
    score of bounding-box predictions to be used by the tracker.
    \item \textbf{Minimum matching tracker IOU}: sets the minimum bounding-box
    IOU (Intersection-Over-Union) that the SORT tracker~\cite{tracking_sort}
    requires to perform obstacle association across frames.
    \item \textbf{Tracking re-initialization frequency}: specifies every how
    many frames detection is run to update the tracker. When detection is not
    run, the SORT tracker falls back on a Kalman filter to linearly extrapolate
    existing bounding box movement. This option bypasses the detection inference
    latency, but quickly incurs significant error.
\end{tightitemize}

The additional \param{} ranges are:
\begin{lstlisting}[language=Python,
    keywordstyle=\color{black},
    commentstyle=\color{codegreen},
    caption={\small \textbf{Values for the other \params{}.}},
    label=a:config-params-full-small]
# Additional Metaparameters
detection-confidence = {0.4, 0.6, 0.7}
minimum-matching-tracker-IOU = {0.1, 0.2, 0.3}
Tracker-re-initialization-frequency = {1, 2, 3}
\end{lstlisting}

\myparagraph{II. The relative contribution of the two \params{} we use} The new
\param{} cartesian product (\Cref{a:config-params-full-small}) is of size
$6*3*3*3*3 = 486$ configurations. Due to computational constraints, we create a
version of the Octopus dataset using 49 training and 23 test scenarios from
the Waymo dataset, with $\tau = 1s$.

\begin{table*}[tb]
    \scriptsize
    \centering
    \caption{\small \textbf{\Param{} contribution to the opportunity gap.} The
    top row shows the score of the global-best policy where no \param{} changes.
    Each row below shows the score of the optimal dynamic policy as one
    additional \param{} is allowed to change. The incremental score increase is
    shown on the right.}
    \input{tables/LAT_AR-other-metaparams.tex}
    \label{t:other-metaparams}
\end{table*}

\Cref{t:other-metaparams} shows the relative contribution of each \param{} to
the score opportunity gap (\cref{ss:opportunity-gap}) between the baseline
global best policy and the optimal dynamic policy. The first two \params{}
(detection model and tracking maximum age) are used in our study and achieve
$83\%$ ($(30.1 - 24.0) / (31.3 - 24.0)$) of the opportunity gap conferred by
using all 5 \params{}. We leave investigation into optimizing the rest of the
\params{} for future work.

%% file: tables/LAT_AR-other-metaparams.tex
\begin{tabular}{@{}lcc@{}}
    \toprule
    \thead{\textbf{Dynamic}           \\ \textbf{\params{}}} & \thead{\textbf{\tmota{}} \\ \textbf{score}} & \thead{\textbf{Score} \\ \textbf{increase}} \\ \midrule
    None (global best policy)                                & 24.0                                        & -                                           \\
    +Detection Model                                         & 28.3                                        & 4.3                                         \\
    +Tracking maximum age                                    & 30.1                                        & 1.8                                         \\
    +Tracking re-init frequency                              & 30.5                                        & 0.4                                         \\
    +Tracking minimum IOU                                    & 31.0                                        & 0.5                                         \\
    +Detection confidence threshold                          & 31.3                                        & 0.3                                         \\ \bottomrule
\end{tabular}

%% file: appendix-training-details.tex
During training, we exclude video segments with no ground truth labels, where
the \mota{} score~\cite{milan2016mot16} is undefined. Then, at test time, we
impute the global static configuration computed over the training dataset.
We clip the regression targets to [-100, 100] (i.e., $\epsilon=100$ in
\cref{eq:final_loss}).
In our evaluation setup we consider an IOU of 0.4 between a prediction and
ground truth to be a true positive. We train the models and evaluate using this
schema.
We use a single GPU (instead of 3 used in Waymo) for evaluation on the
Argoverse dataset, because the high framerate would require over 7 GPUs to
support executing detection in parallel on every frame.

%% file: appendix-hyperparameters.tex
The training hyperparameters used in training the regression and classification
models are shown in \Cref{t:hyperparams}.

\begin{table*}[ht]
  \scriptsize
  \centering
  \caption{\small \textbf{Training Hyperparameters.}}
  \input{tables/LAT_AR-hyperparams.tex}

  \label{t:hyperparams}
\end{table*}

The rest of the hyperparameters are the default used in Scikit-Learn v0.23.2.

%% file: tables/LAT_AR-hyperparams.tex
{
\begin{tabular}{@{}l|cccc@{}}
    \toprule
    \textbf{Method}              & \thead{\textbf{Max Tree} \\ \textbf{Depth}} & \thead{\textbf{Max \#}\\\textbf{of Features}} & \thead{\textbf{\# of} \\ \textbf{Estimators}} & \thead{\textbf{Min Impurity} \\ \textbf{Decrease}} \\ \midrule
    Regression                   & 20                                          & 18                                            & 400                                           &                                           0.000186 \\
    Classification (Joint)       & 8                                           & 3                                             & 400                                           &                                           0.000285 \\
    Classification (Independent) & 7                                           & 4                                             & 200                                           &                                           0.000529 \\
\end{tabular}
}

%% file: appendix-mota-motp.tex
In this work, we demonstrate that environment context can be leveraged to
perform test-time optimization of tracking in streaming settings. Unlike other
perception tasks (e.g. detection~\cite{padilla2021comparative}), where a single
optimization metric is commonly used, in multi-object tracking there is no
consensus on the best metric~\cite{dendorfer2020mot20,luiten2021hota}.
Therefore, in this study we chose to optimize MOTA~\cite{milan2016mot16}, as it
is the metric that most closely aligns with human perception of tracking
quality~\cite{leal2017tracking}.

\begin{figure*}[tbh]
    \begin{minipage}[c]{0.5\textwidth}
        \includegraphics[width=\textwidth]{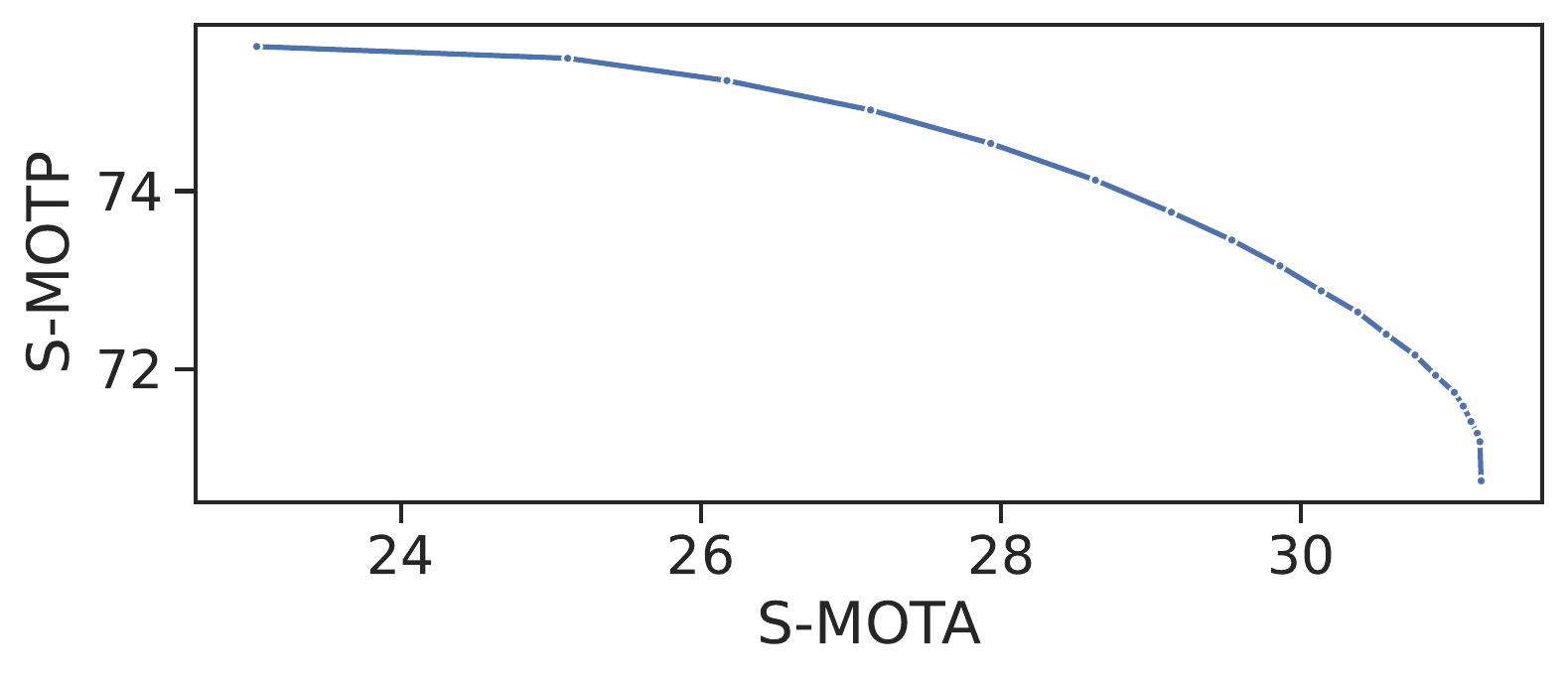}
    \end{minipage}\hfill
    \begin{minipage}[c]{0.47\textwidth}
        \caption{\small \textbf{The pareto frontier of optimal \tmota{} vs.
                \tmotp{} optimizing policies.} The \tmota{} (x-axis) and
            \tmotp{} (y-axis) of optimal policies with gradually varying
            weights (blue curve) from \tmotp{} to \tmota{}.}
        \label{f:mota-motp-pareto-timely}
    \end{minipage}
\end{figure*}

Nevertheless, a battery of other tracking metrics is presented in the
evaluation~\cref{ss:eval-main-results}. The main results (\Cref{t:main-results})
show that the \tmota{}-optimal policy deteriorates in \tmotp{}, and that our
learned policy does the same in the Argoverse dataset. This occurs because
\mota{} and \motp{} are designed to describe fundamentally different properties
in tracking~\cite{milan2016mot16}: Whereas \mota{} equally balances precision,
recall and identification, \motp{} only focuses on precise obstacle
localization. Indeed,~\Cref{t:mota-motp-results} empirically confirms the
conflict between these metrics. The \tmotp{}-optimal policy achieves a far lower
\tmota{} score (23.0) than a policy that optimizes directly for \tmota{} (31.2),
and vice versa for \tmotp{} (75.6 and 71.0, respectively). The conflict between
the \tmota{} and \tmotp{} scores can be illustrated as a pareto frontier
in~\Cref{f:mota-motp-pareto-timely} (generated by linearly interpolating these
metrics).

\begin{figure*}[tbh]
    \begin{minipage}[c]{0.5\textwidth}
        \includegraphics[width=\textwidth]{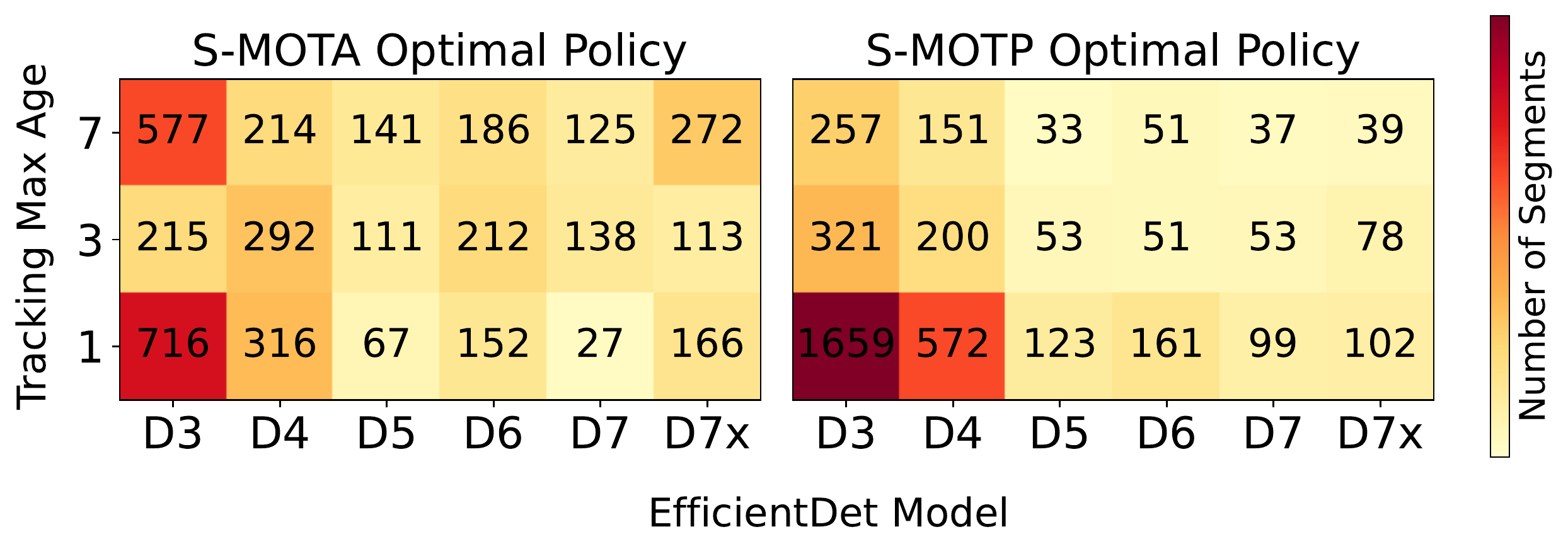}
    \end{minipage}\hfill
    \begin{minipage}[c]{0.47\textwidth}
        \caption{\small \textbf{Different policy strategies are needed to
            optimize \tmota{} and \tmotp{}.} The configuration choice frequency
            (color intensity) of \tmota{}-optimal policy (left) and of the
            \tmotp-optimal policy (right).}
        \label{f:mota-motp-optimal-heatmap}
    \end{minipage}
\end{figure*}

To further visualize the different strategies needed to maximize each metric, we
compare the decision frequency of the \tmota{}-optimal and \tmotp{}-optimal
policies (\Cref{f:mota-motp-optimal-heatmap}). As shown, the \tmotp{}-optimal
policy concentrates on the fastest-models (D3, D4) and on a max-age value of 1,
whereas the \tmota{}-optimal policy's decisions are much more spread out. This
takes place because these models maximize predicted obstacle localization
precision by minimizing prediction lag after the moving ground-truth. The low
max-age minimizes error from SORT's Kalman filter (data not shown). Using weaker
models with a low max-age, however, incurs a higher rate of false-negatives and
ID-switches, which reduces \tmota{} (\Cref{t:mota-motp-results}). We leave
further investigation for future work.

\begin{table*}[tbh]
    \scriptsize
    \centering
    \caption{\small \textbf{Evaluating dynamic policies that optimize \tmota{} and
        \tmotp{}.}}
    \begin{tabular}{@{}l|ccccc@{}}
        \toprule
        \textbf{Method}  & \textbf{\tmota{}}$\uparrow{}$ & \textbf{\tmotp{}}$\uparrow{}$ & \textbf{\tfp{}}$\downarrow{}$ & \textbf{\tfn{}}$\downarrow{}$ & \textbf{\tid{}\textsubscript{sw}}$\downarrow{}$ \\ \midrule
        \tmota{}-optimal & 31.2                          & 71.0                          & 28907                         & 590847                        & 6997                                            \\
        \tmotp{}-optimal & 23.0                          & 75.6                          & 31468                         & 640261                        & 9424                                            \\
    \end{tabular}
    \label{t:mota-motp-results}
\end{table*}

%% file: appendix-study-case.tex
A video of the study case presented in~\cref{ss:eval-explainability} is attached
with the supplementary material (\texttt{study-case.mp4}). Bounding box and ID
annotations are added, matching the color schema of the line-plot that shows the
\tmota{} scores of the different policies.

The 10-12 second mark in the video illustrates a large difference between the
\system{} policy (green) and the global best policy (black). This difference
occurs because the \system{} policy successfully tracks obstacles adjacent to
the road on the left and even more on the right that the global best
policy does not.

%% file: appendix-tsne-visualization.tex
\begin{figure}
    \centering
    \includegraphics[width=\textwidth]{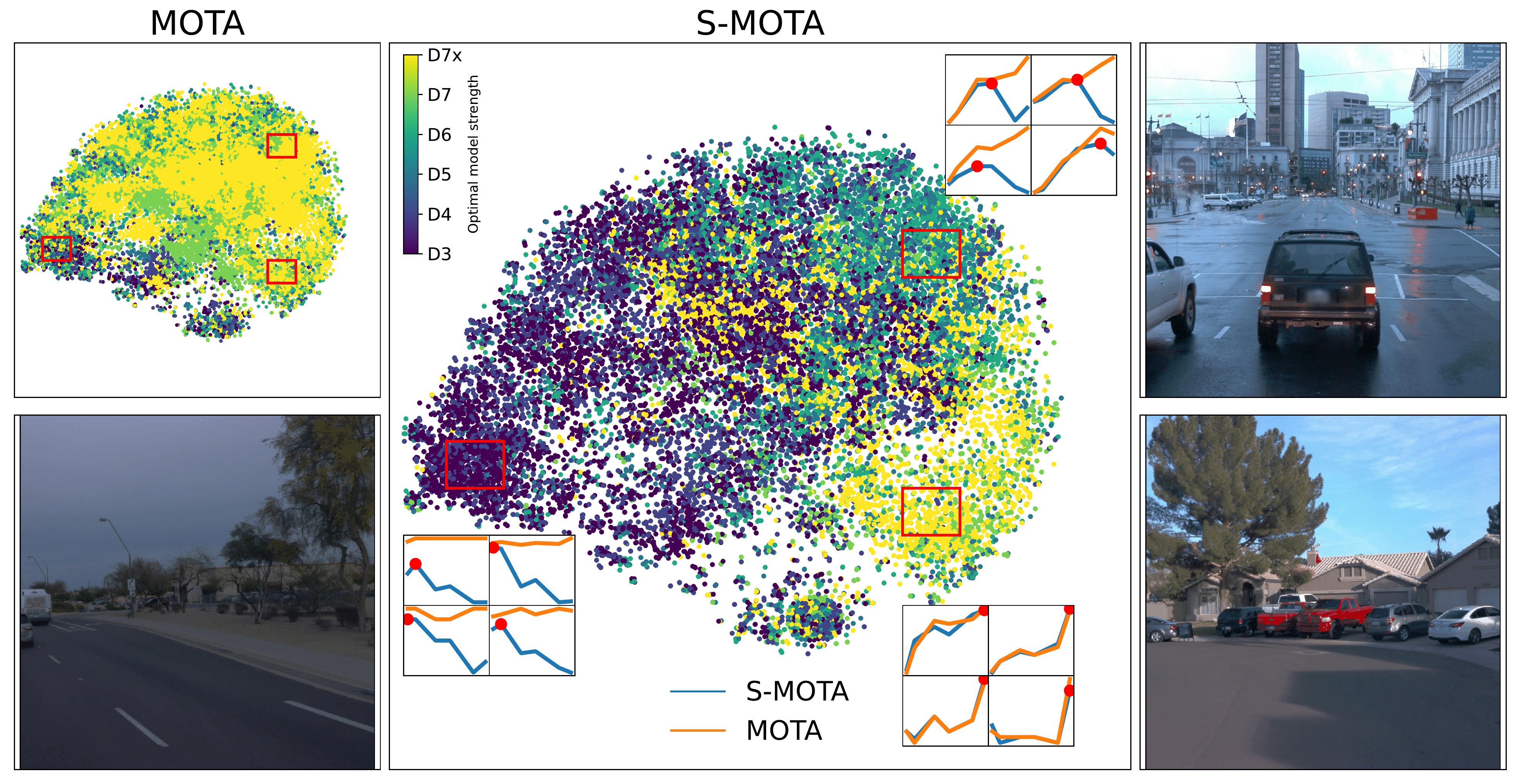}
    \caption{\small \textbf{\mota{} and \tmota{} responses in different regions
    of the t-SNE plot, with video-segment visualization.} This figure
    illustrates the following: \one{} The t-SNE figures presented
    in~\cref{f:tsne-sync-timely} \two{} Three line-plot quartets, illustrating
    the \mota{} and \tmota{} response (y-axis) of nearby points in the figure to
    detection model size (x-axis), \three{} A visualization of a representative
    scenario for each quartet.}
    \label{f:tsne-study-case}
\end{figure}

We present an expanded analysis of the t-SNE plot
in~\cref{ss:eval-explainability}. The figure is presented again
in~\Cref{f:tsne-study-case} with more annotations.

\myparagraph{Proximity in score space}
Three distinct regions in the t-SNE plot are selected and marked with red
rectangles in \Cref{f:tsne-study-case}. Four points, representing four 1-second
video segments, are selected in each region. The \mota{} and \tmota{} scores
(y-axis) of increasing detection model size (x-axis) are plotted for these
scenarios, following the same methodology as \Cref{f:t-mota-sync-mota} and
\Cref{f:example-cases} in \cref{s:introduction}. The y-axis ticks and scale are
omitted to emphasize that nearby video segments have similar,
\textit{normalized}, \mota{} and \tmota{} response to the \params{}. These
scenarios, therefore, also tend to be optimized by the same \param{} values.

\myparagraph{Case studies}
We perform an in-depth analysis of a video segment selected from each of the
three regions indicated in red in~\Cref{f:tsne-study-case}.

\textit{Bottom-left: Turning in difficult conditions.} The ego-vehicle turns
left into a highway. Larger models cannot detect the parked cars in the
background, and therefore do not boost offline \mota{} over smaller models. At
the same time, vehicle turning induces high obstacle-displacement in the frame,
causing accuracy deterioration from higher inference latency. This scene is both
very difficult (minimal offline \mota{} boost from bigger models) and fast (high
accuracy deterioration).

\textit{Bottom-right: Slow movement towards partially-occluded vehicles.}
Stronger models better detect the partially-occluded parked vehicles. The slow
ego-vehicle movement towards non-moving obstacles induces minimal
obstacle-displacement and therefore negligible accuracy degradation.

\textit{Top-right: A mix of still and moving obstacles.} The vehicle is
standing at an intersection. Still or parked cars on the road are mixed with
fast-moving pedestrians. The intermediate-size models, EfficientDet-D5 and D6,
detect most of the still obstacles and are able to keep up with the fast-moving
pedestrians. Larger models (EfficientDet-D7 and D7x), however, introduce too
much inference runtime delay in tracking the moving pedestrians, and sustain
more severe accuracy deterioration as a result.

Taken together, these examples illustrate that the environment context can be
used to infer \mota{} and \tmota{} response to the \params{}. This then enables
test-time \tmota{} optimization by dynamically tuning the \params{}.

%% file: appendix-centroid-viz.tex
\begin{figure*}[tb]
    \centering
    \includegraphics[width=0.9\columnwidth]{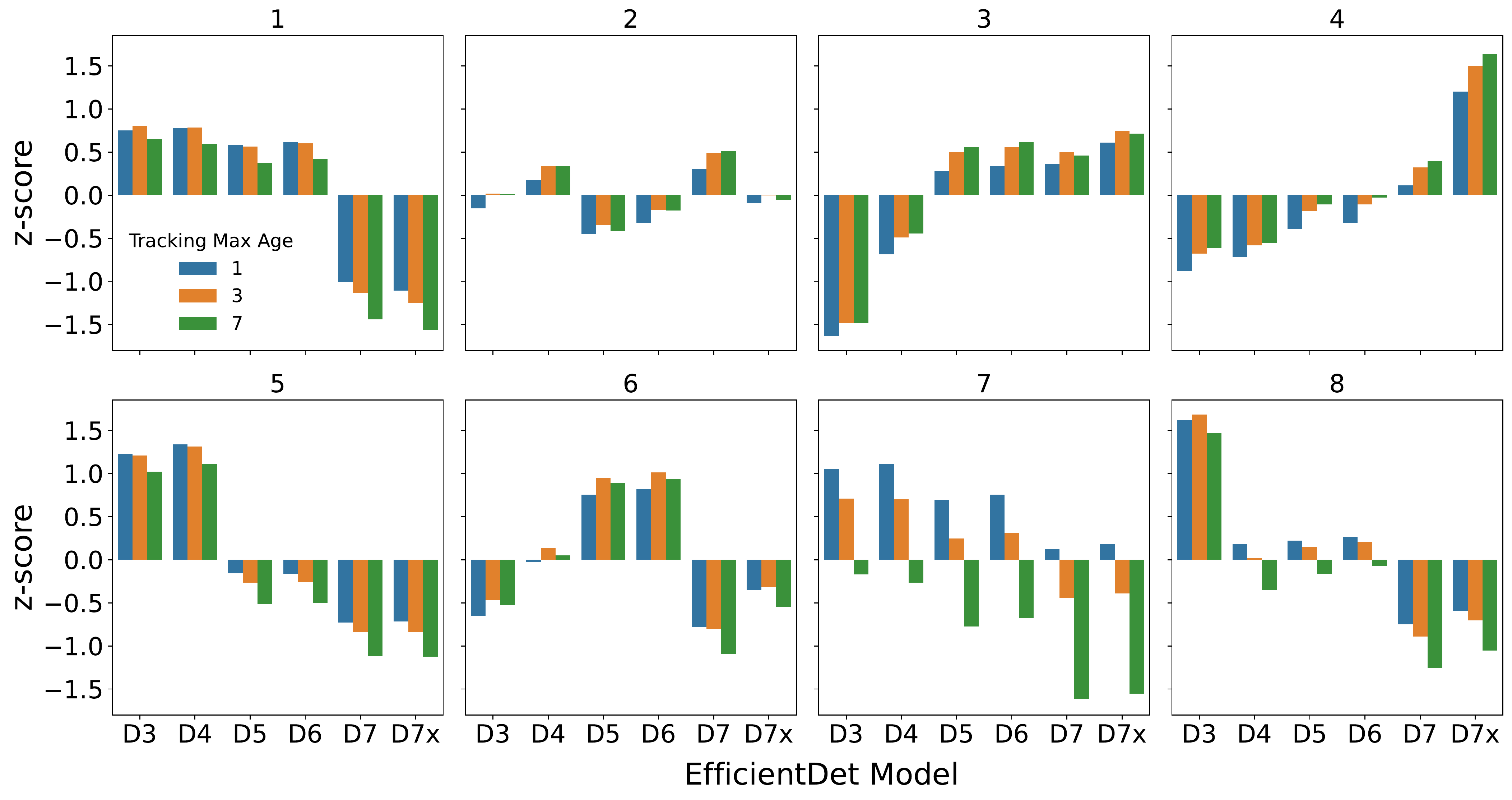}
    \caption{\small \textbf{All eight centroids in the clustering analysis
            performed in \cref{ss:eval-explainability}.}}
    \label{f:full-centroid}
\end{figure*}

\myparagraph{Full score space clustering analysis}
The clustering analysis discussed in~\cref{ss:eval-explainability}
(\Cref{f:score-space}) is extended to all eight centroids, shown
in~\Cref{f:full-centroid}. Several clusters present similar, though slightly
shifted behavior, to the ones presented in the body of the paper. For example,
cluster centroids 1, 5, and 8 represent environments with varying \tmota{}
response to increasing latency. This demonstrates that the performance penalty
(degradation) may occur at different stages as the model size increases.
Clusters 3, 4, and 6 show similar improvement in \tmota{} with larger models,
though in cluster 6, the performance drops for the two largest models, i.e.,
EfficientDet-D7 and D7x. These models' predictions are evaluated 3 frames away
from the ground truth, whereas the rest of the models are evaluated up to 2
frames away. This may raise the possibility that the increase in frame gap from
the ground truth causes a more significant degradation to \tmota{} in cluster 4
if it contains faster moving scenes than cluster 6.

\myparagraph{Cluster scenario visualization}
Curated videos of several of the clusters' video segments are attached to the
supplementary material (\texttt{cluster-\{4, 7, 8\}.mp4}) The cluster numbers
correspond to the centroid number plotted in~\cref{f:full-centroid}.

When comparing clusters 4 and 8, we observe a clear trend: \one{} video segments
with a preference for larger models (cluster 4) show more subdued obstacle
displacement from the ego vehicle camera's perspective and \two{} the opposite
in video segments with preference for smaller models (cluster 8).

In cluster 7, a lower tracking maximum age has better performance. Here, we
observe video segments with two primary modes of obstacle behavior: \one{}
occlusions or obstacles leaving the scene, and \two{} very fast obstacle
movement, where the IoU of the obstacle to its previous location is small,
causing SORT~\cite{tracking_sort} to do an ID switch.
In both cases, maintaining the old tracklets for a shorter amount of time
reduces false positives and improves performance.

These observations illustrate the idea that visible properties of the
environment context may be leveraged to discern the mode of \param{} score
behavior. These properties are used as features in \system{} to optimize
\tmota{} at test time.

%% file: evaluation-ablations-B.tex

We conduct an ablation study of the experimental setup and the design choices.

\myparagraph{Baseline subtraction}
As a variance reduction technique, we consider regressing over the relative
score improvement from $h^{global}$ instead of absolute scores as described in
\ref{ss:policy-regression}.
Thus, we avoid dedicating model capacity to learn environment ``difficulty''
properties, which apply to all configurations in a manner that does not change
their score order.
As this distinction is not relevant for predicting configuration order, it only
introduces noise to the downstream ranking task~\cite{yogatama2014efficient}.
Accordingly, in \Cref{t:class-ablations}, we observe a $0.3$ point decrease
in \tmota{} score on the Waymo dataset when baseline subtraction is ablated.

\myparagraph{Classification \emph{vs.} regression}
%
We compare the regress-then-rank approach described in
\cref{ss:policy-regression} against the classification formulation (using Random
Forests).
This approach directly predicts the \param{} values of the best configuration
from the environment features using $m$ classifiers for each of the $m$
\params{}.
In \Cref{t:class-ablations} we show the resulting score in "classification
(joint)", indicating a reduced score compared to regression with baseline
subtraction.

%

\myparagraph{Independent \emph{vs.} joint \param{} optimization} We examine
the degree of independence between \param{} choices in the optimization
process.
To this end, we consider a new setting "classification (independent)" in
\Cref{t:class-ablations}. This is done by separately learning to predict the
value of each \param{}, while holding the other \param{} values constant (in
essence assuming convexity).
We show that joint classification performs $0.4$ \tmota{} points better than
independent. 

\begin{table}[tbh]
  \scriptsize
  \centering
  \caption{\small \textbf{Comparing configuration ranking approaches}}
  \input{tables/LAT_AR-classification-ablations.tex}
  \label{t:class-ablations}
\end{table}

%% file: tables/LAT_AR-classification-ablations.tex
{
\begin{tabular}{@{}l|cccccc@{}}
    \toprule
    \textbf{Method}                     & \textbf{\tmota{}}$\uparrow{}$ & \textbf{\tmotp{}}$\uparrow{}$ & \textbf{\tfn{}}$\downarrow{}$ & \textbf{\tfp{}}$\downarrow{}$ & \textbf{\tid{}\textsubscript{sw}}$\downarrow{}$ \\ \midrule
    Regression w/ baseline subtraction  & 27.9                          & 72.3                          & 31489                         & 608870                        & 8966                                            \\
    Regression w/o baseline subtraction & 27.6                          & 72.5                          & 31405                         & 610617                        & 8982                                            \\
    Classification (independent)        & 27.2                          & 72.9                          & 32725                         & 615537                        & 9326                                            \\
    Classification (joint)              & 27.6                          & 72.7                          & 35107                         & 611196                        & 8829                                            \\
\end{tabular}
}

%% file: appendix-policy-neural-network.tex
The key idea in this study is that the environment context can be leveraged to
optimize streaming tracking accuracy at test time. The Octopus policy model
presented is implemented using engineered features and random forest
regression. A natural question is whether the policy performance can be
improved using deep neural networks. To this end, we evaluated policy design
that incorporates a conventional, convolutional neural network
(CNN)~\cite{he2016deep} as well. We found that performance is on par with the
global best policy, much worse than the solution using engineered features
(see~\Cref{t:neural-network}).
In order to get good performance, we believe that the model needs to predict
features at the granularity of instance-level motion (e.g.
instance-flow~\cite{center_track}) because the tracking score that is being
predicted is defined at this granularity. We believe that the model does not
get good score because it does not capture these features well. Further
investigation is needed.

\subsection{Methodology}
We use a ResNet50 backbone~\cite{he2016deep} that takes as input the middle
frame of each 1-second video segment and classifies the best configuration using
a new MLP. The backbone is pretrained using QDTrack~\cite{qdtrack} on the
BDD100k dataset~\cite{yu2018bdd100k} (QDTrack's zero-shot accuracy on Waymo is
40.3, on par with the optimal policy score in offline settings). The backbone
output is average-pooled and fed to an MLP of width of 256 with one hidden
layer. The MLP output is fed into separate linear layers that generate the
logits for each \param{} in order to separately classify the values of the
optimal configuration. This follows the methodology described
in~\cref{app:eval-ablations-B}. The model is trained using the cross-entropy
loss on the Waymo dataset, using the methodology described
in~\cref{ss:eval-setup}. The logits are initialized with a small bias so that
the model behaves like the global best policy at the start of training. The
model is trained using AdamW~\cite{loshchilov2017decoupled} with a learning rate
of 1e-4 and weight decay of 0.01 for 10 epochs.

\subsection{Results}

The results are shown in \Cref{t:neural-network}. The CNN achieves negligible
performance improvement over the global best policy.

\begin{table}[tbh]
    \scriptsize
    \centering
    \caption{\small\textbf{Neural-network based policy performance}}
    \input{tables/LAT_AR-neural-network.tex}

    \label{t:neural-network}
  \end{table}

%% file: tables/LAT_AR-neural-network.tex
{
\begin{tabular}{@{}l|cccccc@{}}
    \toprule
    \textbf{Method}                    & \textbf{\tmota{}}$\uparrow{}$ & \textbf{\tmotp{}}$\uparrow{}$ & \textbf{\tfp{}}$\downarrow{}$ & \textbf{\tfn{}}$\downarrow{}$ & \textbf{\tid{}\textsubscript{sw}}$\downarrow{}$ \\ \midrule
    Global best                        & 25.1                          & 72.2                          & 33616                         & 633159                        & 11212                                           \\ \midrule
    Neural network based policy        & 25.2                          & 72.3                          & 45658                         & 611827                        & 8056                                            \\
\end{tabular}
}

%% file: appendix-faster-hardware-simulation.tex
Hardware performance is expected to improve over time. We therefore evaluate
\system{} on faster hardware execution by simulating 50\% faster inference of
the detection model measured on the V100 GPU. We repeat the evaluation on the
Argoverse dataset described in~\cref{ss:eval-main-results}, showing the results
in~\Cref{t:argo-latency-simulation}. The results show that the Octopus policy
with closed-loop prediction outperforms the global best static policy by 2.8
\tmota{}, up from the 1.7 \tmota{} result using the V100 GPU latency readings.
\begin{table}[tbh]
    \scriptsize
    \centering
    \caption{\small\textbf{Performance on Argoverse with 50\% faster GPU inference}}
    \input{tables/LAT_AR-argo-simulated-latencies.tex}
    \label{t:argo-latency-simulation}
\end{table}


%% file: tables/LAT_AR-argo-simulated-latencies.tex
\begin{tabular}{@{}l|ccccc@{}}
    \toprule
    \textbf{Method}                         & \textbf{\tmota{}}$\uparrow{}$ & \textbf{\tmotp{}}$\uparrow{}$ & \textbf{\tfp{}}$\downarrow{}$ & \textbf{\tfn{}}$\downarrow{}$ & \textbf{\tid{}\textsubscript{sw}}$\downarrow{}$ \\ \midrule
    Global best                             & 54.5                          & 77.3                          & 8508                          & 44965                         & 1173                                            \\ \midrule
    Optimal                                 & 63.7                          & 75.5                          & 6897                          & 36634                         & 649                                             \\
    Optimal from the prev. segment          & 57.6                          & 75.5                          & 9655                          & 40800                         & 776                                             \\ \midrule
    \textbf{\system{}} with:                &                               &                               &                               &                               &                                                 \\
    Ground truth from current segment       & 59.3                          & 75.9                          & 7951                          & 40405                         & 815                                             \\
    Ground truth from prev. segment         & 57.9                          & 76.3                          & 7988                          & 41553                         & 958                                             \\
    \textbf{Prediction from prev. segment } & \textbf{57.3}                 & \textbf{76.4}                 & \textbf{7911}                 & \textbf{42489}                & \textbf{958}                                    \\ \bottomrule
\end{tabular}